\documentclass[twocolumn]{style}

\usepackage{amsmath,amssymb,mathtools}
\usepackage{tabularx}
\usepackage{booktabs}
\usepackage{colortbl}
\usepackage{cleveref}
\usepackage{fontawesome5}
\usepackage{inconsolata}
\usepackage{xspace}
\usepackage{xurl}
\usepackage{fontawesome5}

\newcommand{\methodname}{\mbox{GeoMem-VLA}\xspace}
\newcommand{\geotok}{GeoPlan\xspace}
\newcommand{\papertitle}{\methodname: Internalizing 3D Scene Understanding into Temporal Memory for Vision-Language-Action Models}
\title{\rmfamily\papertitle}

\newcommand{\obs}{\mathbf{o}}
\newcommand{\acts}{\mathbf{a}}
\newcommand{\state}{\mathbf{s}}
\newcommand{\lang}{\ell}
\newcommand{\gtok}{\mathbf{c}}
\newcommand{\mem}{\mathcal{M}}
\newcommand{\pts}{\mathbf{P}}
\newcommand{\eef}{\mathbf{W}}
\newcommand{\policy}{\pi_\theta}

\DeclareTextFontCommand{\textbf}{\bfseries}
\DeclareRobustCommand{\corrmark}{\text{\faEnvelope[regular]}}
\DeclareRobustCommand{\hficon}{\raisebox{-0.15em}{\includegraphics[height=1em]{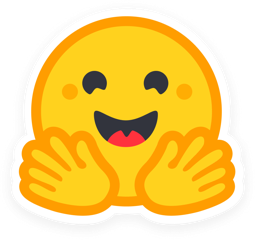}}}

\ifdefined\fi
\setcitestyle{numbers,square,citesep={,}}

\AtBeginDocument{%
  \crefname{table}{Table}{Tables}%
  \Crefname{table}{Table}{Tables}%
  \crefname{figure}{Figure}{Figures}%
  \Crefname{figure}{Figure}{Figures}%
  \crefname{section}{Section}{Sections}%
  \Crefname{section}{Section}{Sections}%
  \crefname{appendix}{Appendix}{Appendices}%
  \Crefname{appendix}{Appendix}{Appendices}%
  \crefname{subappendix}{Appendix}{Appendices}%
  \Crefname{subappendix}{Appendix}{Appendices}%
}

\hypersetup{
  pdftitle={GeoMem-VLA: Internalizing 3D Scene Understanding into Temporal Memory for Vision-Language-Action Models},
  pdfauthor={Qinfan Zhang, Yifan Luo, Haoyuan Shi, Guanlin Liu, Han Dong, Haozhe Shan, Yong Dai, Xiancong Ren, Xiaozhu Ju}
}

\renewcommand{\authorlist}{%
  \authorfont\rmfamily\mdseries
  Qinfan Zhang$^{1,3,\dag}$, Yifan Luo$^{2,3,\dag}$, Haoyuan Shi$^{3,4}$,
  Guanlin Liu$^{3,6}$, Han Dong$^{3}$,\\[0.4em]
  Haozhe Shan$^{5}$, Yong Dai$^{3}$, Xiancong Ren$^{3,\ddag}$,
  Xiaozhu Ju$^{3,\corrmark}$%
}

\renewcommand{\affiliationfont}{\fontsize{8.4}{10}\selectfont}
\renewcommand{\affiliationlist}{%
  \affiliationformat[1]{Beihang University}\,\textperiodcentered\,
  \affiliationformat[2]{Renmin University}\,\textperiodcentered\,
  \affiliationformat[3]{Beijing Innovation Center of Humanoid Robotics}\\[1pt]
  \affiliationformat[4]{USTC}\,\textperiodcentered\,
  \affiliationformat[5]{Fudan University}\,\textperiodcentered\,
  \affiliationformat[6]{The Chinese University of Hong Kong, Shenzhen}%
}

\contribution[\dag]{Core Contributors}
\contribution[\ddag]{Project Leader}
\contribution[\corrmark]{Corresponding Author}

\abstract{
Vision-Language-Action (VLA) models typically predict actions from current 2D observations, whereas existing 3D policies infer geometry per step without retaining it across an episode. Action generation under occlusion and across multiple stages requires a compact spatial abstraction that persists over time without reconstructing a full 3D scene at each step.
Based on this insight, we introduce \textbf{\methodname}, a framework that internalizes explicit 3D scene and motion supervision into learnable latent \geotok tokens, leaving the vision-language backbone's native visual embeddings unaligned. During training, a frozen geometry teacher and demonstration trajectories supervise these tokens via an auxiliary flow-matching head over current scene point maps and future end-effector tracks. Across steps, the policy stores observation and \geotok representations in separate memory banks, merges similar adjacent states into event memories, and fuses retrieved history with current features to predict actions.
At inference, the 3D teacher and auxiliary head are removed, and action generation relies solely on the latent \geotok tokens and episode memory without explicit scene reconstruction. \methodname achieves average success rates of 98.9\% on LIBERO, 91.10\% and 91.24\% on RoboTwin 2.0 Clean and Randomized settings, and 78.0\% across five real-world manipulation tasks.
}
\checkdata[\faDesktop~Website]{\url{https://geomem-vla.github.io}}
\checkdata[\faGithub~Code]{\url{https://github.com/GeoMem-VLA/GeoMem-VLA}}
\checkdata[\hficon~HuggingFace]{\url{https://huggingface.co/rxc205/GEOMEM-VLA}}

\fancypagestyle{firststyle}{%
  \fancyhf{}%
  \fancyhfoffset[L,R]{0pt}%
  \fancyhead[L]{\vskip 3mm
    \includegraphics[height=10mm]{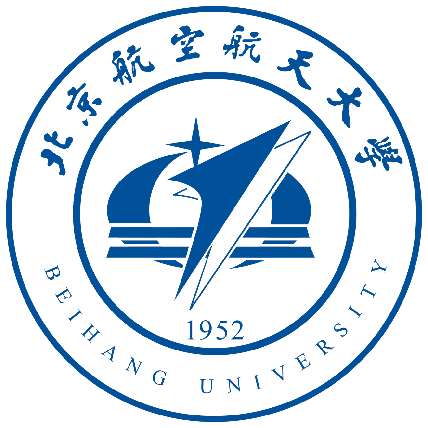}%
    \hspace{5mm}%
    \includegraphics[trim=346bp 34bp 346bp 34bp,clip,height=10mm]{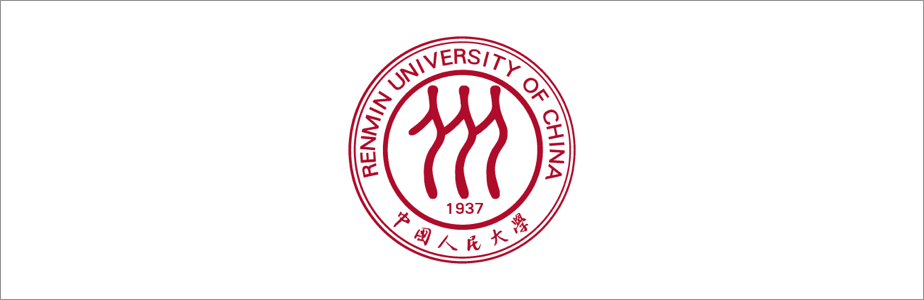}%
  }%
  \fancyhead[R]{\vskip 3mm\includegraphics[height=11.5mm]{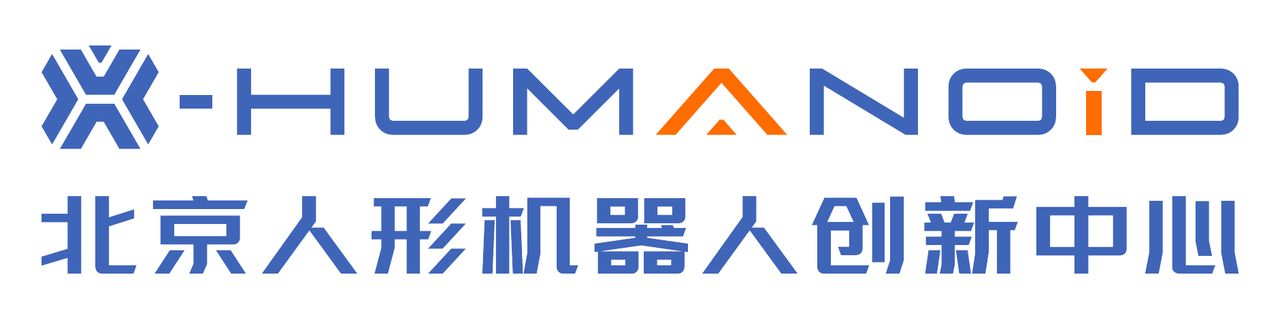}}%
  \fancyfoot[C]{\thepage}%
}

\definecolor{frontlinkblue}{HTML}{3977AC}

\begin{document}
\hypersetup{urlcolor=frontlinkblue}
\maketitle
\hypersetup{urlcolor=spatheme}
\raggedbottom

\section{Introduction}
\begin{figure*}[t]
  \centering
  \includegraphics[width=\textwidth]{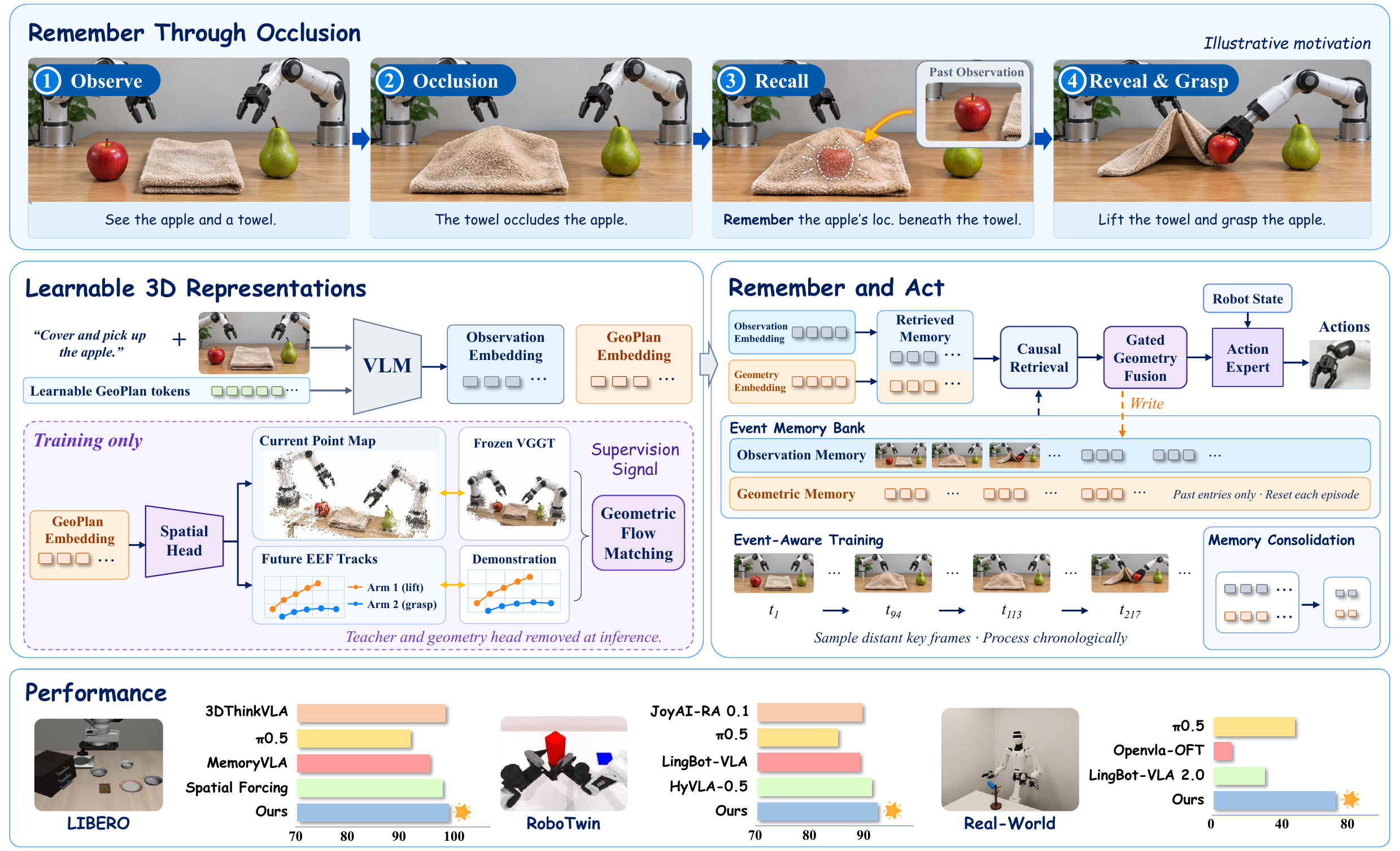}
  \caption{
 \textbf{Overview of \methodname.}
\textbf{Top:} In the illustrative cover and pick up the apple task, the current observation alone may not reveal the apple’s location once it is occluded by the towel, motivating the retention of spatial context over time. \textbf{Middle:} \methodname learns geometry-supervised GeoPlan representations and retains them in causal episode memory for action generation. \textbf{Bottom: } Performance comparisons across manipulation benchmarks and real-world tasks.}
  \label{fig:teaser}
\end{figure*}

\label{sec:intro}
Building on advances in large language models, vision-language-action
(VLA) models have attracted attention for language-conditioned manipulation
\citep{rt2,openvla,pi0,VLA-Trace}. Most VLA models use pretrained vision-language models (VLMs)
to inherit the visual and semantic understanding capabilities and predict actions through tokenized
prediction or dedicated action experts. However, as shown in \Cref{fig:teaser}, the policy that relies on current
2D observations offers limited support
for precise 3D spatial reasoning and decision-making that depend on
interaction history.

Recent work on 3D-aware VLAs has explored how to leverage the rich geometric priors of the physical environment \citep{spatialforcing, geopredict}.
Such supervision allows explicit geometric targets to shape policy representations.
Memory-augmented
policies, meanwhile, retain perceptual and task context from earlier
observations \citep{memoryvla}.
Connecting these directions raises a central question:
\textbf{How can geometric supervision shape spatial memory that guides action?}
This places the spatial content of memory at the center of
policy representation learning.

We propose \methodname, a VLA framework that couples geometry-supervised
\geotok representations with causal episode-level memory. During training,
learnable \geotok tokens summarize the multimodal context and condition an
auxiliary flow-matching head that predicts current-scene point maps and
future Cartesian end-effector tracks. A frozen VGGT teacher \citep{vggt}
provides scene-geometry targets, while demonstration trajectories provide
future motion supervision. These complementary targets train the tokens to
encode scene structure and task-related motion. At inference, the teacher
and reconstruction head are removed from the action-generation path;
the learned tokens provide compact geometric context for the policy to generate action.

To preserve this context, we maintain geometric memory alongside
perceptual observation memory. Current representations retrieve
relevant preceding features, and adaptive fusion integrates historical
context before action generation. Similarity-based consolidation compresses redundant entries while retaining feature-distinct temporal states. We further use event-aware
training that samples temporally separated observations from the same
episode while preserving their chronological order and causal access.
This exposes retrieval to a broader span of interaction history than
contexts formed only from neighboring frames. Together, geometric
supervision and temporal retrieval connect what the policy learns about
the scene with what it can retain for later decisions.

Our evaluation examines spatial understanding, long-horizon decision-making,
and applicability across manipulation tasks. On LIBERO and RoboTwin~2.0,
\methodname achieves $98.9\%$ average success across the four LIBERO suites
and $91.10\%$ / $91.24\%$ on RoboTwin Clean / Randomized. Complementary
real-world protocols target geometric alignment, long-horizon interaction
and non-markovian decision process, connecting the evaluation to the spatial and temporal capability of the architecture.

Our contributions are summarized as follows:

\begin{itemize}
  \item \textbf{Geometry-Supervised Spatial Representations.}
  We learn compact \geotok tokens from current-scene point maps and future
  end-effector tracks, providing geometric conditioning without explicit
  reconstruction during action inference.
  \item \textbf{Geometry-Aware Event Memory.}
  We couple geometric representations with causally temporal retrieval and
  train event-aware memory to retain  key information alongside
  perceptual and task context.
  \item \textbf{Evaluation Across Manipulation Settings.}
  Experiments of geometry-supervised memory across diverse tasks through LIBERO, RoboTwin and real-world manipulation tasks examine geometric precision and the use of historical context in long-horizon manipulation.
\end{itemize}

\section{Related Work}
\label{sec:related}
\paragraph{\textbf{Vision-Language-Action Models.}}
Advances in VLMs \citep{liu2024improved,karamcheti2024prismatic}
have enabled VLAs through robot-data scaling, vision-language pretraining,
and expressive action decoders. RT-1 \citep{rt1} scales multitask learning
on demonstrations; RT-2 \citep{rt2} and OpenVLA \citep{openvla} transfer
pretrained knowledge to discrete action prediction. $\pi_0$ \citep{pi0},
CogACT \citep{cogact}, and GR00T \citep{gr00t} use continuous action
experts, while $\pi_{0.5}$ \citep{pi05} adds heterogeneous data and
semantic subtask prediction for open-world generalization. FAST
\citep{fast} compresses action sequences via frequency-domain
tokenization; OpenVLA-OFT \citep{openvlaoft} uses parallel decoding and
action chunking for efficient execution. StarVLA \citep{starvla}
provides a modular backbone--expert framework. These advances leave
open how task-relevant 3D structure should be encoded and retained
across decisions.

\paragraph{\textbf{3D Representations and Geometric Supervision.}}
Spatially informed VLAs use depth \citep{li2026bridgevla,fan2026any3d},
feature-level geometric supervision \citep{abouzeid2025geoaware},
and predictive representations. Spatial Forcing \citep{spatialforcing}
aligns intermediate visual features with frozen VGGT features;
3DThinkVLA \citep{3dthinkvla} combines feature alignment with spatial
reasoning distillation. Other methods model motion and scene evolution
\citep{bhat20253d,li20253ds}: GeoPredict \citep{geopredict} couples
future robot-keypoint trajectories with predictive Gaussian geometry,
while GaussianDream \citep{gaussiandream} learns a latent prefix through
current and future Gaussian scene prediction. These approaches enrich
spatial representations, but geometric supervision alone does not
specify how the resulting information should be maintained and accessed
throughout an episode.

\paragraph{\textbf{Temporal and Spatial Memory.}}
History-aware policies use recurrent states, visual prompts, and
retrievable memory. RoboFlamingo \citep{roboflamingo} aggregates
sequential features with a recurrent policy head; TraceVLA
\citep{tracevla} encodes tracked motion as visual traces. HAMLET
\citep{hamlet} aggregates distinctive moment tokens in a lightweight
memory module. MemoryVLA \citep{memoryvla} utilizes historical information by storing and retrieving perceptual and cognitive history in a memory bank; EventVLA
\citep{eventvla} predicts keyframes to preserve sparse visual evidence. BridgeVLA++ \citep{bridgevlapp} combines temporal observation history with explicit point-cloud spatial memory. These designs trade retained detail against temporal coverage and cost, but how scene and motion
targets shape latent history for later decisions remains open. We focus
on jointly learning and using spatial representations across time to
guide manipulation.
\label{sec:related:positioning}

\section{Method}
\label{sec:method}
\subsection{Preliminaries}
\label{sec:method:prelim}
\begin{figure*}[t]
  \centering
  \includegraphics[width=\textwidth]{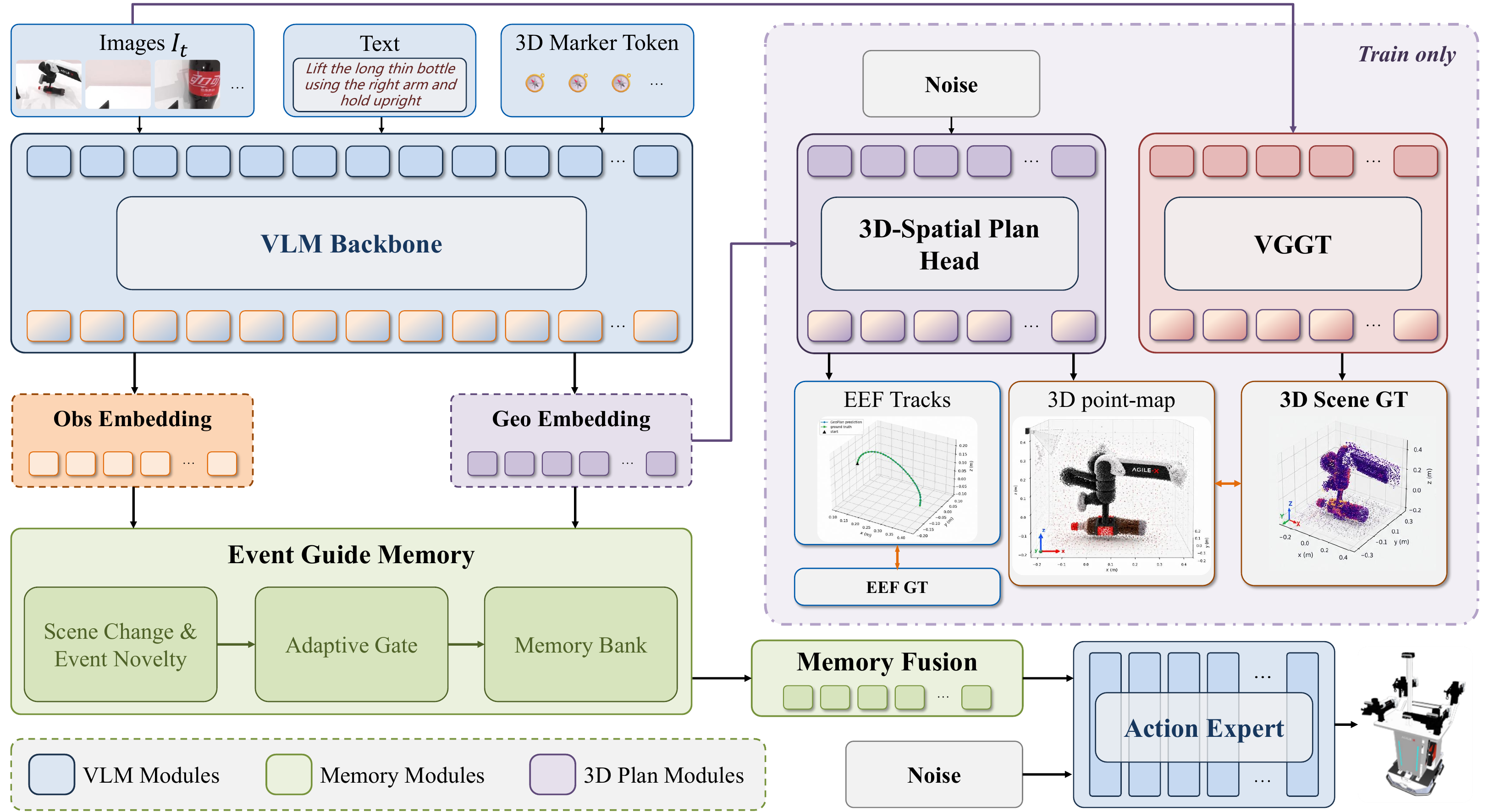}
  \caption{\textbf{Architecture of \methodname.}
  Images, language, and learnable 3D marker tokens are encoded by the VLM. The geometry branch is trained with scene-geometry targets and demonstration-derived future end-effector tracks, whereas the action expert is trained to predict robot actions. Geometric and observational features are retained in episodic memory and fused to form the context for action generation. The geometry teacher and reconstruction head are used only during training.
  }
  \label{fig:overview}
\end{figure*}

\paragraph{Vision-Language-Action Models.}
We formulate the language-conditioned manipulation as sequential decision making.
At control step $t$, the policy receives multi-view RGB observations
$\obs_t=\{I_t^{(v)}\}_{v=1}^{V}$, proprioceptive state
$\state_t\in\mathbb{R}^{d_s}$, and a language instruction $\lang$, and
predicts a continuous action chunk
\begin{equation}
 \begin{aligned}
 \widehat{\mathbf{A}}_t &=\policy(\obs_t,\state_t,\lang)\\
 &=(\widehat{\acts}_t,\ldots,\widehat{\acts}_{t+H-1})
 \in\mathbb{R}^{H\times d_a}.
 \end{aligned}
 \label{eq:vla_policy}
\end{equation}
Here $H$ is the prediction horizon, while $d_a$ and the semantics of each
action are determined by the robot embodiment and control interface.

\paragraph{GR00T-Style Action Expert.}
We instantiate a GR00T-style flow-matching action expert \citep{gr00t}
in StarVLA \citep{starvla}, using Qwen3-VL-4B \citep{qwen3vl} as the
backbone and a conditional Diffusion Transformer as the expert. Let
$\mathbf{A}_t=(\acts_t,\ldots,\acts_{t+H-1})
\in\mathbb{R}^{H\times d_a}$ be a normalized demonstration chunk and
$\mathbf{C}_t$ the vision-language context. For
$\boldsymbol{\epsilon}^{a}\sim\mathcal{N}(\mathbf{0},\mathbf{I})$ and
$\rho\sim p_{\mathrm{act}}$ from the configured Beta-derived schedule,
define $\mathbf{A}_{t,\rho}
=(1-\rho)\boldsymbol{\epsilon}^{a}+\rho\mathbf{A}_t$.
The expert minimizes the noise-to-data velocity error:
\begin{equation}
\begin{aligned}
\mathcal{L}_{\mathrm{action}}
 &=\mathbb{E}_{\mathbf{A}_t,\rho,\boldsymbol{\epsilon}^{a}}
 \left[\frac{\|\mathbf{e}^{a}_t\|_F^2}{H d_a}\right],\\
\mathbf{e}^{a}_t
 &=v_{\psi}^{\mathrm{act}}(\mathbf{A}_{t,\rho},\rho;\mathbf{C}_t,\state_t)
 -(\mathbf{A}_t-\boldsymbol{\epsilon}^{a}).
\end{aligned}
\label{eq:action_fm}
\end{equation}
At inference, Euler integration transforms Gaussian noise into an
action chunk.

\paragraph{VGGT Geometry Teacher.}
Visual Geometry Grounded Transformer (VGGT) is a feed-forward geometry model that predicts camera parameters and
dense 3D attributes, including depth maps, point maps, and point tracks, from
one or more RGB images \citep{vggt}. Its transformer alternates frame-wise and
global self-attention before applying task-specific prediction heads.

\subsection{Geometry-Supervised Latent Representations}
\label{sec:method:overview}
\label{sec:method:bridge}

\Cref{fig:overview} shows the overview of \methodname. We omit the batch dimension throughout and use $D_v$ for the VLM hidden
dimension and $D_g$ for the geometric working dimension.
Furthermore, we append $K$ \geotok markers to the vision-language sequence. Let
$\mathbf{q}_{1:K}=[\mathbf{q}_1;\ldots;\mathbf{q}_K]$ denote their input
embeddings, with $\mathbf{q}_i\in\mathbb{R}^{D_v}$. Their final-layer hidden
states are extracted and projected to the geometric working space:
\begin{equation}
  \begin{aligned}
  \mathbf{z}_t &=\mathrm{VLM}([\obs_t;\lang;\mathbf{q}_{1:K}])_{\mathrm{GeoPlan}},\\
  \gtok_t &=\mathrm{Proj}(\mathbf{z}_t)\in\mathbb{R}^{K\times D_g}.
  \end{aligned}
  \label{eq:inseq}
\end{equation}
Here $\mathrm{VLM}$ denotes the pretrained vision-language backbone.
The subscript $\mathrm{GeoPlan}$ selects the $K$ marker
positions, giving $\mathbf{z}_t\in\mathbb{R}^{K\times D_v}$.
$\mathrm{Proj}$ applies layer normalization and a learned linear map from
$D_v$ to $D_g$ to each output token.
The projected tokens $\gtok_t$ both
condition the auxiliary geometry head and form the geometric memory entries.

\paragraph{Supervision Targets.}
\label{sec:method:teacher}
The frozen VGGT teacher's current-view point map is spatially
pooled to a $P\times P$ grid to form the scene pseudo-target,
with the corresponding confidence scores weighting the scene
supervision loss. In parallel, we construct a \emph{future EEF
trajectory} from demonstrations, consisting of $n_e$ temporally
ordered Cartesian end-effector positions over the action horizon.
When only joint states are available, these positions are
recovered through forward kinematics.

To ensure geometric consistency between the scene and motion
targets, we align them in a shared coordinate frame and scale.
During offline preprocessing, we obtain camera extrinsics and
metric depth references by replaying training demonstrations,
and estimate the teacher's scale from corresponding depth
measurements. The future EEF positions are then transformed
into the VGGT point-map coordinate frame. Both targets are
normalized using the centroid and root-mean-square (RMS)
radius of the scale-aligned point map
(\cref{app:eef_alignment}).
This represents the intended EEF motion relative to the
observed scene geometry.

Finally, we concatenate the normalized point-map and
EEF-trajectory targets:
\begin{equation}
  \mathbf{Y}_t=[\pts_t;\eef_t]
  \in\mathbb{R}^{(P^2+n_e)\times3},
  \label{eq:target}
\end{equation}
where $\pts_t\in\mathbb{R}^{P^2\times3}$ denotes the scene
point-map target and $\eef_t\in\mathbb{R}^{n_e\times3}$
denotes the future EEF trajectory target.

\paragraph{Auxiliary Flow Matching.}
\label{sec:method:geohead}
Let $\epsilon$ be a noise tensor with the same shape as $\mathbf{Y}_t$ and
independent standard Gaussian entries, and let $\tau\sim\mathcal{U}[0,1]$
be the flow time, distinct from control step $t$. Define
$\mathbf{Y}_{t,\tau}=(1-\tau)\epsilon+\tau\mathbf{Y}_t$.
A modality-aware transformer $v_\phi$, with learned parameters $\phi$,
predicts the velocity from this noisy target, flow time, and geometric
condition $\gtok_t$. Its output has the same shape as $\mathbf{Y}_t$:
\begin{equation}
\begin{aligned}
\mathbf{E}_t
 &=v_\phi(\mathbf{Y}_{t,\tau},\tau,\gtok_t)-(\mathbf{Y}_t-\epsilon),\\
\mathcal{L}_{\mathrm{geo}}
 &=\sum_{m\in\{\mathrm{scene,eef}\}}\lambda_m
 \mathbb{E}_{\tau,\epsilon}\!
 \left[\|\mathbf{E}_{t,m}\|_{w_m}^{2}\right].
\end{aligned}
\label{eq:geoloss}
\end{equation}
Here $m$ indexes the Scene or EEF segment, and the subscript $m$ selects that segment from the
predicted and target velocities.
The action objective remains the action expert’s flow-matching loss:
\begin{equation}
  \mathcal{L}=\mathcal{L}_{\mathrm{action}}+\alpha\mathcal{L}_{\mathrm{geo}}.
  \label{eq:total}
\end{equation}
Here $\mathcal{L}_{\mathrm{action}}$ is the action flow-matching objective in
Eq.~\eqref{eq:action_fm}, and $\alpha$ controls geometric supervision relative to
action learning.
Model dimensions and optimization settings are given in \cref{app:hyperparams}.

\subsection{Geometry Event-Aware Memory}
\label{sec:method:memory}
At each control step, the policy retrieves history within each representation
stream, updates its bounded episode memory, and then injects retrieved geometry
into the retrieved observation representation. We maintain two observation
streams and one geometric stream, indexed by
$j\in\{\mathrm{obs},\mathrm{geo}\}$.
The observation streams $\mathbf{x}^{\mathrm{obs}}_t\in\mathbb{R}^{N_o\times D_v}$ contain resized visual tokens
and the final
non-padding VLM state token.
While the geometric stream
$\mathbf{x}^{\mathrm{geo}}_t=\gtok_t\in\mathbb{R}^{K\times D_g}$ contains
geometry-supervised tokens. The first stream jointly form observation
memory; the second forms geometric memory.

\paragraph{Causal Temporal Retrieval.}
Let $\mem^j_{<t}$ contain only earlier entries from the same episode.
Each stream retrieves from its own bank:
\begin{equation}
  \widetilde{\mathbf{x}}^j_t
  =R_j(\mathbf{x}^j_t,\mem^j_{<t};t).
  \label{eq:memory_retrieval}
\end{equation}
$R_j$ uses current tokens as queries and stored tokens as keys and values,
adds learned timestamp encodings to the keys, and fuses the retrieved
context with the current representation. Point-map- and future-motion-
supervised GeoPlan states are written to a dedicated bank, where similar
adjacent entries are consolidated. Denoting the retrieved geometric
tokens by $\widetilde{\gtok}_t
=\widetilde{\mathbf{x}}^{\mathrm{geo}}_t$, a gate combines them with
retrieved visual tokens; the result and the other observation stream
condition the action expert. Further details appear in \cref{app:geo_fusion}.

\paragraph{Event-Aware Training.}
Retrieval precedes writing, preventing the current entry from retrieving
itself. We then append $(t,\operatorname{sg}(\mathbf{x}^j_t))$ to bank $j$,
where $\operatorname{sg}$ stops gradients through stored history. If the
ordered bank $\{(s_r,\mathbf{m}^j_r)\}_{r=1}^{L}$ exceeds its capacity, it
selects the most similar temporally adjacent pair:
\begin{equation}
 r_j^\star=\arg\max_{1\leq r<L}
 \operatorname{cos}\!\left(
   \operatorname{vec}(\mathbf{m}^j_r),
   \operatorname{vec}(\mathbf{m}^j_{r+1})\right).
 \label{eq:memory_consolidation}
\end{equation}
The pair is replaced by its elementwise mean and labeled with the later
timestamp. Repeated updates therefore preferentially compress redundant
states within stable manipulation phases, while feature-distinct transitions
are more likely to remain in separate slots. We call the resulting latent
summaries \emph{event-level key memories}: they summarize temporally distinct
states induced by representation change. Banks are reset at episode boundaries, and deployment memory never
receives future observations or demonstration actions.
The replay diagnostics in \cref{app:cases}
also examine overflow-based consolidation.

\paragraph{Deployment.}
\label{sec:method:objective}
The VGGT teacher and geometry decoder are omitted from action inference. The
backbone still computes the latent geometric tokens, and memory retrieval and similarity-based consolidation remain active. No explicit point-map or EEF-track decoding is required
for action generation; latent-token and memory operations remain part of
the deployed policy.

\section{Experiments}
\label{sec:exp}
\subsection{Experimental Setup}
\label{sec:exp:setup}

\paragraph{Simulation Benchmarks and Evaluation.}
We evaluate on LIBERO \citep{libero}, using four 10-task suites (Spatial, Object, Goal, and Long) covering language-conditioned single-arm manipulation. In addition, we also evaluate RoboTwin 2.0 \citep{robotwin}, which comprises 50 bimanual manipulation tasks in Clean and Randomized settings. Our setup uses 7-dimensional end-effector delta actions with gripper control and prediction horizon $H=8$ for LIBERO, and 14-dimensional joint targets including gripper commands with $H=50$ for RoboTwin. We run 50 trials per LIBERO task and 100 trials per RoboTwin task per setting. At inference time, the task-adaptive execution horizon $E_t \leq H$ determines how many predicted actions are executed before replanning.
Further model, optimization, and memory settings appear in \cref{app:hyperparams}. Published results in
\cref{tab:libero_main,tab:robotwin_main}.

\begin{figure*}[t]
  \centering
  \includegraphics[width=\textwidth]{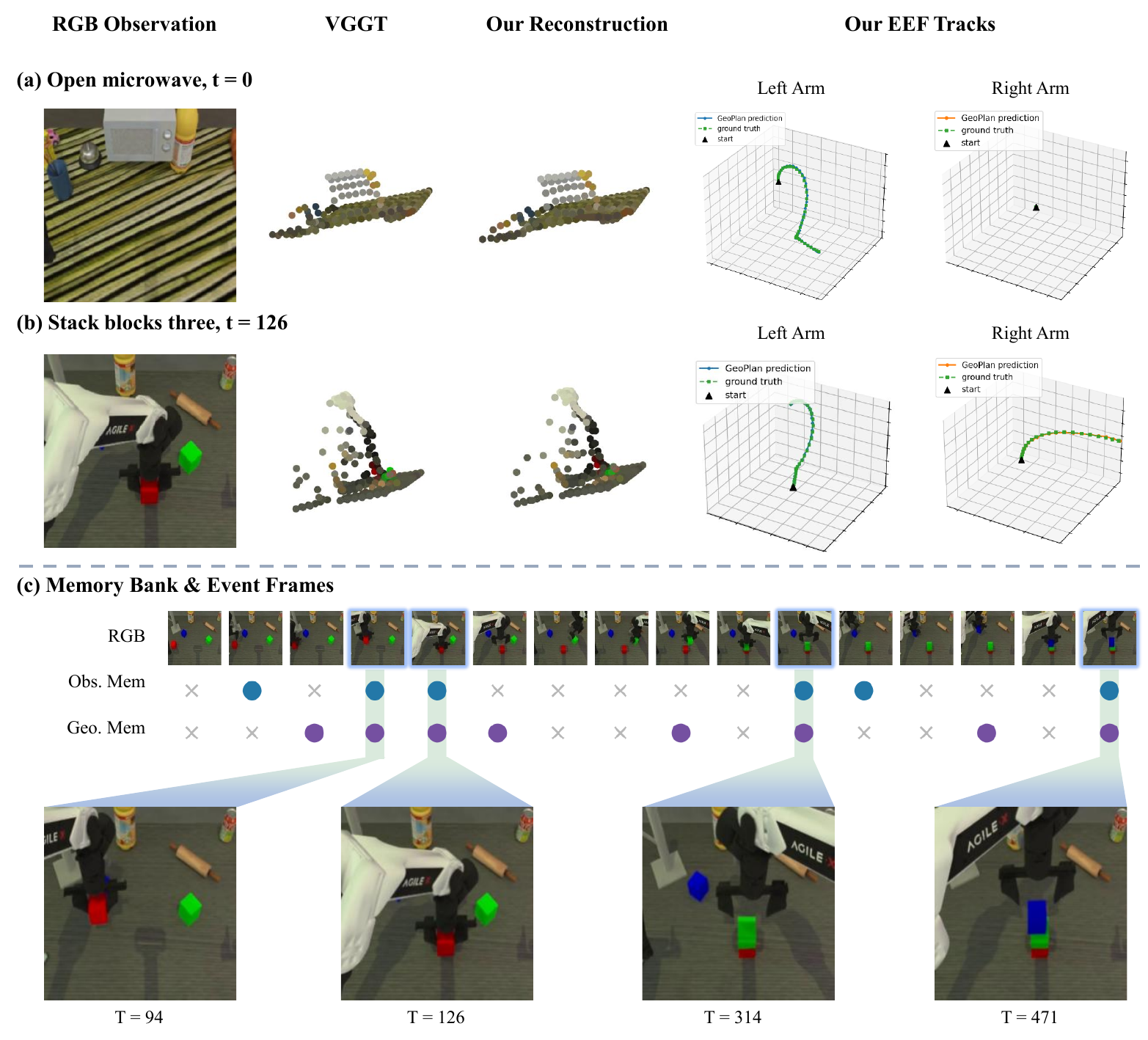}
  \caption{\textbf{Geometry, future motion, and retained memory in RoboTwin.} (a,b) Randomized examples show RGB observations, VGGT point-map targets, predicted point maps, and future EEF tracks (solid: prediction; dashed: demonstration). (c) Final memory retention in Randomized stacking: the observational row shows the intersection of both observation banks, while the geometric row shows geometric memory (dots: retained; crosses: not retained; vertical shaded bands: shared timestamps), with corresponding frames enlarged below.
  }
  \label{fig:case_geometry}
\end{figure*}

\begin{figure*}[t]
    \centering
    \includegraphics[width=\linewidth]{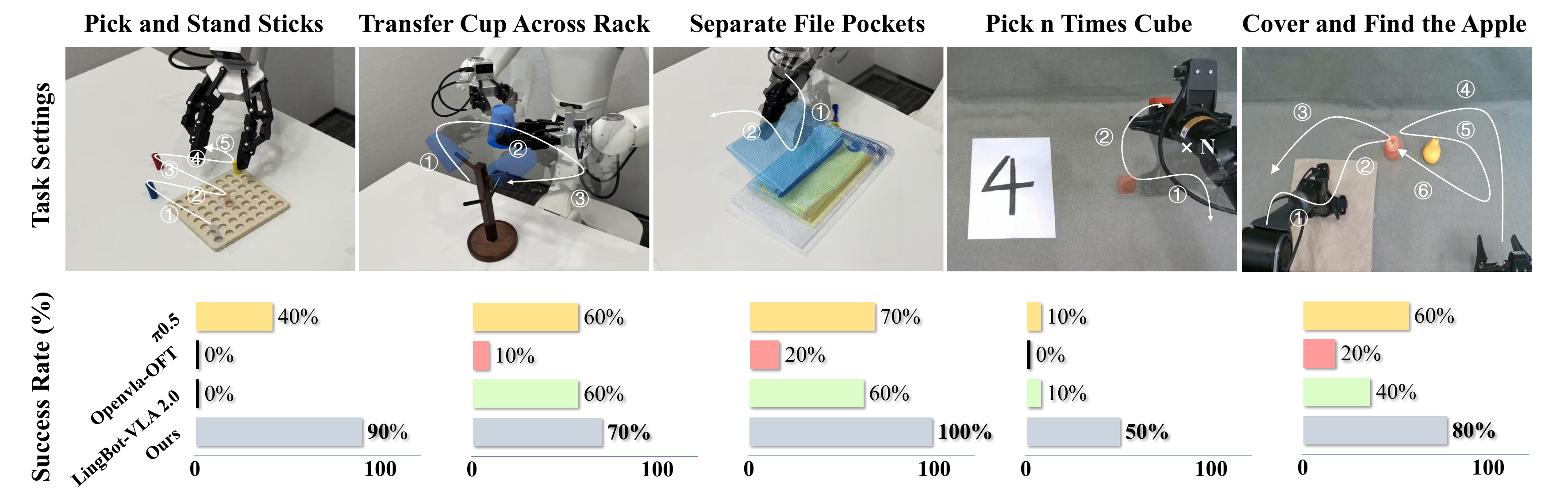}
    \caption{
\textbf{Real-world Manipulation Evaluation of GeoMem-VLA.}
The figure shows the real-world experiments task settings and success rates across five tasks.
Each bar reports the success rate over 10 independent closed-loop rollouts.
    }
    \label{fig:RealWorldExper}
\end{figure*}

\begin{table}[t]
\centering
\caption{\textbf{LIBERO Success Rates (\%).} References beside method names
identify the first-party source for each published result. Averages cover
the four displayed suites.
Bold and underline mark the
highest and second-highest distinct values in each column, including ties.}
\label{tab:libero_main}
\small
\setlength{\tabcolsep}{2.5pt}
\begin{tabularx}{\linewidth}{@{}l*{5}{>{\centering\arraybackslash}X}@{}}
\toprule
Method & Spat. & Obj. & Goal & Long & Avg. \\
\midrule
OpenVLA (FT) \citep{openvla} & 84.7 & 88.4 & 79.2 & 53.7 & 76.5 \\
TraceVLA \citep{tracevla} & 84.6 & 85.2 & 75.1 & 54.1 & 74.8 \\
SpatialVLA \citep{spatialvla} & 88.2 & 89.9 & 78.6 & 55.5 & 78.1 \\
4D-VLA \citep{fourdvla} & 88.9 & 95.2 & 90.9 & 79.1 & 88.6 \\
DreamVLA \citep{dreamvla} & 97.5 & 94.0 & 89.5 & 89.5 & 92.6 \\
$\pi_0$ \citep{pi0} & 96.8 & 98.8 & 95.8 & 85.2 & 94.2 \\
$\pi_{0.5}$ \citep{pi05} & 98.8 & 98.2 & 98.0 & 92.4 & 96.9 \\
GeoPredict \citep{geopredict} & 98.0 & 98.2 & 95.7 & 94.0 & 96.5 \\
MemoryVLA \citep{memoryvla} & 98.4 & 98.4 & 96.4 & 93.4 & 96.7 \\
CronusVLA \citep{cronusvla} & 97.3 & 99.6 & 96.9 & 94.0 & 97.0 \\
OpenVLA-OFT \citep{openvlaoft} & 97.6 & 98.4 & 97.9 & 94.5 & 97.1 \\
GeoVLA \citep{geovla} & 98.4 & 99.0 & 96.6 & \textbf{96.6} & 97.7 \\
AVA-VLA \citep{avavla} & 99.2 & 99.6 & 97.9 & \underline{96.2} & 98.2 \\
Spatial Forcing \citep{spatialforcing} & 99.4 & 99.6 & \underline{98.8} & 96.0 & 98.5 \\
3DThinkVLA \citep{3dthinkvla} & \textbf{100.0} & \textbf{100.0} & \underline{98.8} & 95.8 & \underline{98.7} \\
\midrule
\rowcolor{gray!15}
\textbf{\methodname} & \underline{99.6} & \underline{99.8} &
\textbf{99.6} & \textbf{96.6} & \textbf{98.9} \\
\bottomrule
\end{tabularx}
\end{table}

\begin{table}[t]
\centering
\caption{\textbf{RoboTwin 2.0 success rates (\%).} Clean and Randomized
refer to the 50-task evaluation (called Easy/Hard in some reports).
Boldface and underlining indicate the highest and second-highest distinct values in each column, including ties.}
\label{tab:robotwin_main}
\small
\setlength{\tabcolsep}{3pt}
\begin{tabularx}{\linewidth}{@{}l*{3}{>{\centering\arraybackslash}X}@{}}
\toprule
Method & Clean & Rand. & Avg. \\
\midrule
X-VLA \citep{zheng2026x} & 70.00 & 39.00 & 54.50 \\
Track4Action \citep{track4action} & 80.44 & 81.48 & 80.96 \\
ABot-M0 \citep{abotm0} & 86.06 & 85.08 & 85.57 \\
Qwen-VLA-Instruct \citep{qwenvla} & 86.10 & 87.20 & 86.65 \\
LingBot-VLA \citep{lingbotvla} & 88.56 & 86.68 & 87.62 \\
StarVLA-OFT \citep{starvla} & 88.18 & 88.32 & 88.25 \\
StarVLA-GR00T \citep{starvla} & 88.00 & 88.50 & 88.25 \\
JoyAI-RA 0.1 \citep{joyaira} & 90.48 & 89.28 & 89.88 \\
HyVLA-0.5 \citep{hyvla} & \underline{90.90} & \underline{90.10} & \underline{90.50} \\
\midrule
\rowcolor{gray!15}
\textbf{\methodname} & \textbf{91.10} & \textbf{91.24} & \textbf{91.17} \\
\bottomrule
\end{tabularx}
\end{table}

\subsection{Main Results}
\label{sec:exp:main}

\Cref{tab:libero_main} shows that \methodname achieves the highest displayed
LIBERO average ($98.9\%$), $0.4$ points above Spatial Forcing. Spatial,
Object, and Goal are near ceiling ($99.6\%$--$99.8\%$); Long is more
discriminative, where our $96.6\%$ ties the highest displayed score and
exceeds MemoryVLA by $3.2$ points. The advantage is therefore not confined
to the three near-saturated suites.

On RoboTwin, \methodname reaches $91.10\%$ on Clean and $91.24\%$ on
Randomized, averaging $91.17\%$ (\cref{tab:robotwin_main}). It exceeds
HyVLA-0.5 by $0.20$ and $1.14$ points in the two settings, respectively,
and the closest published StarVLA-GR00T reference by $2.92$ points on
average. The results show strong performance across both end-effector and
joint-action benchmarks, including randomized evaluation. Since published
methods may differ in training data, inputs, and protocols, these comparisons
provide context rather than a controlled ranking; the component ablations
below assess the contributions of geometry and memory.
\begin{table}[t]
\centering
\caption{\textbf{Component, supervision, and history ablations on LIBERO.}
Success rates (\%); 50 trials per task. EEF-only and Scene-only use only
EEF-trajectory or scene-point-map supervision, respectively.
Settings are detailed in \cref{sec:exp:ablation}.}
\label{tab:ablation_libero}

\small
\setlength{\tabcolsep}{2pt}
\definecolor{ablationblue}{RGB}{235,239,255}

\begin{tabular*}{\linewidth}{@{\extracolsep{\fill}}lccccc@{}}
\toprule
Configuration & Spat. & Obj. & Goal & Long & Avg. \\
\midrule

\rowcolor{ablationblue}
\multicolumn{6}{c}{\textbf{Full model}} \\
\textbf{\methodname\ (full)}
& \textbf{99.6} & \textbf{99.8} & \textbf{99.6}
& \textbf{96.6} & \textbf{98.9} \\

\midrule
\rowcolor{ablationblue}
\multicolumn{6}{c}{\textbf{Component ablations}} \\
w/o temporal memory
& 99.4 & 99.6 & 99.4 & 94.2 & 98.2 \\
w/o geometry branch
& 98.0 & 99.6 & 98.8 & 95.2 & 97.9 \\
w/o geo. \& w/o memo.
& 97.8 & 98.8 & 97.4 & 92.0 & 96.5 \\

\midrule
\rowcolor{ablationblue}
\multicolumn{6}{c}{\textbf{Geometric supervision}} \\
Scene-only
& 98.2 & 99.6 & 99.4 & 95.8 & 98.3 \\
EEF-only
& 98.6 & 99.2 & 99.0 & 95.4 & 98.1 \\

\midrule
\rowcolor{ablationblue}
\multicolumn{6}{c}{\textbf{History retrieval}} \\
w/o observation history
& 99.0 & 99.8 & 99.0 & 93.2 & 97.8 \\
w/o GeoPlan history
& 98.8 & 99.6 & 99.2 & 92.8 & 97.6 \\

\bottomrule
\end{tabular*}
\end{table}

\begin{table}[t]
\centering
\caption{\textbf{Component ablations on RoboTwin 2.0.}
Success rates (\%) across 50 tasks, with 100 episodes per task and setting.
Average is the mean of Clean and Randomized.}
\label{tab:ablation_robotwin}

\small
\setlength{\tabcolsep}{2pt}
\definecolor{ablationblue}{RGB}{235,239,255}

\begin{tabular*}{\linewidth}{@{\extracolsep{\fill}}lccc@{}}
\toprule
Configuration & Clean & Rand. & Avg. \\
\midrule

\rowcolor{ablationblue}
\multicolumn{4}{c}{\textbf{Full model}} \\
\textbf{\methodname\ (full)}
& \textbf{91.10} & \textbf{91.24} & \textbf{91.17} \\

\midrule
\rowcolor{ablationblue}
\multicolumn{4}{c}{\textbf{Component ablations}} \\
w/o temporal memory
& 89.84 & 90.52 & 90.18 \\
w/o geometry branch
& 89.68 & 89.74 & 89.71 \\
w/o geo. \& w/o memo.
& 88.00 & 88.50 & 88.25 \\

\bottomrule
\end{tabular*}
\end{table}

\subsection{Ablation Study}
\label{sec:exp:ablation}

\paragraph{Geometry and Memory.}
\Cref{tab:ablation_libero,tab:ablation_robotwin} compare the full model
with two component-removal variants. The full model stores \geotok tokens in a
geometric bank, consolidates similar adjacent entries, and gates retrieved
geometry into the observation context. Removing GeoPlan eliminates its auxiliary
supervision, geometric tokens, geometric memory bank, and gated fusion,
while retaining the observation banks. Removing temporal
memory retains current geometric tokens and supervision but disables
historical retrieval from all banks.

On LIBERO, the observation-memory variant lowers average success from
$98.9\%$ to $97.9\%$, with drops of $1.6$ points on Spatial and $1.4$
on Long. Without temporal memory, the average falls to $98.2\%$; Long
falls from $96.6\%$ to $94.2\%$, compared with $0.2$-point drops on each
other suite. On RoboTwin, removing GeoPlan lowers the Clean/Randomized
mean from $91.17\%$ to $89.71\%$, while removing memory yields $90.18\%$;
both interventions reduce success in Clean and Randomized. The full model
exceeds these variants on 35 and 27 of 50 RoboTwin tasks, respectively,
after averaging each task across settings (\cref{tab:robotwin_full}).
These results show both components impact the model's task ability.
Furthermore, geometric capabilities are particularly effective in multi-stage tasks. Historical GeoPlan information has additional control value when observation history is still available, and current geometric representations have not been able to fully replace this historical pathway.

\paragraph{Scene and EEF Supervision.}
EEF-only and Scene-only achieve $98.1\%$ and $98.3\%$ average success
(\cref{tab:ablation_libero}), exceeding StarVLA-GR00T ($96.5\%$) and the
variant without the geometry branch ($97.9\%$). On Spatial, they reach
$98.6\%$ and $98.2\%$, versus $98.0\%$ without geometry. Joint supervision
further improves Spatial success to $99.6\%$, gaining $1.0$ and $1.4$
points over the two single-target variants, and performs best across all
four suites with a $98.9\%$ average. These results support combining
scene-structure reconstruction and future-motion prediction to learn
geometric representations for action.

\paragraph{Observation and GeoPlan History.}
To examine the contributions of the two memory streams, we train two
history-ablation models separately. The variant without GeoPlan history
disables retrieval from the geometric bank during both training and
evaluation, while retaining current GeoPlan tokens and observation-history
retrieval. The variant without observation history disables retrieval from
both observation banks during training and evaluation, while retaining
geometric-history retrieval. Both variants retain current observations,
current GeoPlan tokens, and geometric fusion. The full model and both
variants use aligned EEF supervision (\cref{app:eef_alignment}).

As shown in \cref{tab:ablation_libero}, removing GeoPlan history yields
$98.8\%$, $99.6\%$, $99.2\%$, and $92.8\%$ on Spatial, Object, Goal, and
Long, respectively, the average success is $97.6\%$. Compared with the full model,
average success decreases by $1.3$ percentage points and Long suite success by
$3.8$ points. Removing Observation history yields $97.8\%$ average success
and $93.2\%$ on Long suite, corresponding to decreases of $1.1$ and $3.4$ points.
Spatial, Object, and Goal remain near ceiling under both
ablations, whereas the largest drops occur on Long.
Both single-history variants also underperform the variant
without temporal memory on Long suite ($94.2\%$).
These variants fuse a history-conditioned representation
from one stream with current-step features from the other.
This asymmetric temporal context may produce inconsistent
cues about task progress, particularly at subtask transitions.
The full model retrieves both histories and achieves the
best Long performance among these configurations ($96.6\%$).

\subsection{Representation and Memory Diagnostics}
\label{sec:exp:analysis}
To examine how geometry-supervised GeoPlan representations connect current spatial understanding to temporal memory, we replay an episode of
\texttt{open\_microwave} and \texttt{stack\_blocks\_three} in the Clean and
Randomized RoboTwin settings with the 200k step checkpoint. For each episode, we decode point maps and
future EEF tracks at four frames and process 16 uniformly sampled frames
causally through memory. This is an offline demonstration diagnostic. Additional details are provided in \cref{app:cases}.

\Cref{fig:case_geometry}(a,b) and \cref{tab:case_coverage} examine the spatial and motion content of current GeoPlan states: the auxiliary head reconstructs VGGT point-map targets and demonstration-derived EEF tracks, with mean point-map RMSE of 0.064–0.094 in normalized units and mean EEF RMSE of 3.07–3.74 mm across the four cases. The same GeoPlan states are written as entries in geometric memory. \cref{fig:case_geometry}(c) shows entries retained across the initial layout and later stacking stages; the retrieval visualization in \cref{fig:robotwincases}(d) shows access to earlier entries. These diagnostics follow the latent interface from current prediction to temporal storage and retrieval, while \Cref{tab:ablation_libero,tab:ablation_robotwin} evaluate the assembled pathways through closed-loop task success.

\subsection{Real-World Experiments}
\label{sec:exp:real}

\paragraph{Experimental setup.}
We conduct real-world experiments to evaluate \methodname in manipulation scenarios involving geometric reasoning and temporal context.
Specifically, we design five tasks with distinct spatial and sequential requirements as shown in \Cref{fig:RealWorldExper}: \textit{pick and stand sticks}, \textit{transfer cup across the rack}, \textit{separate file pockets} on the Astribot S1 platform and \textit{pick n times cube}, \textit{cover and find the apple under towel} on the AgileX platform.
Astribot S1 is a humanoid robot equipped with two 7-DoF arms, while AgileX Cobot Magic 2 provides two 6-DoF manipulation arms with parallel grippers.
We collect 60 demonstrations per task and \methodname is compared to representative VLA baselines, including LingBot-VLA 2.0 \citep{lingbotvla2}, $\pi_{0.5}$ \citep{pi05}, and OpenVLA-OFT \citep{openvlaoft}, under the same evaluation protocol.
During evaluation, we conduct 10 independent rollouts for each task under randomized object configurations and report the success rate.

\paragraph{Result analysis.}
\Cref{fig:RealWorldExper} shows that \methodname achieves robust performance across five real-world manipulation tasks.
Specifically, for comprehensive long-horizon manipulation tasks, including \textit{pick and stand sticks}, \textit{transfer a cup across the rack}, and \textit{cover and find the apple}, \methodname achieves 90\%, 70\%, and 80\% success rates.
These tasks require both precise spatial interaction and maintaining task-relevant context throughout multi-stage execution, such as in \textit{pick and stand sticks} task, the robot needs to repeatedly identify individual sticks, grasp them from a shared workspace, and place them into a precise upright configuration.
The results show that the geometry-supervised representations provide spatially grounded features for reliable grasping and placement, while the historical geometric context effectively maintains task progress.
Moreover, for the spatially precise \textit{separate file pockets} task, \methodname achieves 100\% success rate by accurately distinguishing and manipulating the target file pocket among cluttered stacked configurations.
For \textit{pick n times cube}, the challenge is maintaining progress over repeated interactions. The policy needs to remember how many successful grasp-and-place cycles have been completed while adapting to the current scene. The temporal memory mechanism preserves task progress across repeated operations, allowing \methodname to complete the required number of interactions.

\section{Conclusion}
\label{sec:conclusion}
Inspired by the thought patterns of human action, we propose \methodname to illustrate how humans conceptualize and store spatial information of the current scene in their minds as memory to guide action generation during decision-making.
This method allows spatial information learned through reconstruction to remain available as latent context throughout task execution. LIBERO and RoboTwin results support its applicability across manipulation tasks. Experiments on two physical robot platforms further cover precise spatial interaction, changing visibility, multi-stage manipulation, and non-Markov decision making. These results support geometry-supervised latent memory as a practical connection between 3D representation learning and action generation over time, effectively improving VLA model's decision-making ability.


\setlength{\bibsep}{2pt plus .5pt minus .5pt}
\bibliographystyle{plainnat}
\bibliography{custom}

\clearpage
\appendix
\setcounter{figure}{0}
\setcounter{table}{0}
\setcounter{equation}{0}
\renewcommand{\thefigure}{S\arabic{figure}}
\renewcommand{\thetable}{S\arabic{table}}
\renewcommand{\theequation}{S\arabic{equation}}
\renewcommand{\theHfigure}{supp.\arabic{figure}}
\renewcommand{\theHtable}{supp.\arabic{table}}
\renewcommand{\theHequation}{supp.\arabic{equation}}

\section*{Appendix}

\Cref{app:hyperparams} lists the training and inference configurations and details the EEF track coordinate alignment (\cref{app:eef_alignment}). \Cref{app:geo_fusion} describes the geometric memory fusion module, and \cref{app:robotwin_full} reports per-task RoboTwin 2.0 results. \Cref{app:cases} provides details of the representation and memory diagnostics, and the final appendix documents the real-world hardware, tasks, evaluation protocol, and baseline implementation.

\section{Implementation Details and Hyperparameters}
\label{app:hyperparams}
\cref{tab:hyperparams} summarizes the model and training configurations used to obtain the reported results. Settings shared across environments are stated explicitly, while environment-specific settings are listed separately.
\begin{table*}[t]
\caption{\textbf{Training and inference configurations.} All platforms use
Qwen3-VL-4B-Instruct~\citep{qwen3vl} and a frozen VGGT~\citep{vggt} teacher. Learning rates are grouped
by the backbone, action expert, and remaining trainable modules.}
\label{tab:hyperparams}
\centering\small
\begin{tabular*}{\linewidth}{@{\extracolsep{\fill}}lcccc@{}}
\toprule
Setting & LIBERO & RoboTwin & Astribot & AgileX \\
\midrule
Action dimension & 7 & 14 & 16 & 7 \\
State dimension & 8 & 14 & 16 & 7 \\
Prediction horizon $H$ & 8 & 50 & 50 & 50 \\
Training budget (updates) & 300000 & 300000 & 30000 & 30000 \\
Per-device batch size (frames) & 16 & 16 & 16 & 16 \\
Gradient accumulation & 4 & 8 & 1 & 2 \\
Training seed & 42 & 42 & 42 & 42 \\
Backbone learning rate & $10^{-5}$ & $10^{-5}$ & $5\!\times\!10^{-5}$ & $5\!\times\!10^{-5}$ \\
Action-expert learning rate & $10^{-4}$ & $10^{-4}$ & $5\!\times\!10^{-5}$ & $5\!\times\!10^{-5}$ \\
Other-module learning rate & $2.5\!\times\!10^{-5}$ & $2.5\!\times\!10^{-5}$ & $5\!\times\!10^{-5}$ & $5\!\times\!10^{-5}$ \\
Warmup updates & 5000 & 5000 & 1000 & 5000 \\
Schedule after warmup & Cosine & Cosine & Constant & Constant \\
Minimum learning rate & $10^{-6}$ & $10^{-6}$ & --- & --- \\
Weight decay & $10^{-8}$ & $10^{-8}$ & $10^{-10}$ & $10^{-10}$ \\
Teacher input size & 336 & 336 & 336 & 336 \\
Scene loss weight $\lambda_{\mathrm{scene}}$ & 0.5 & 0.5 & 0.5 & 0.5 \\
EEF loss weight $\lambda_{\mathrm{eef}}$ & 3.0 & 3.0 & 3.0 & 3.0 \\
Geometry loss coefficient $\alpha$ & 0.1 & 0.1 & 0.1 & 0.1 \\
Memory capacity & 16 & 8 & 16 & 16 \\
Sample group size & 32 & 16 & 32 & 32 \\
Euler integration steps & 4 & 16 & 4 & 16 \\
\bottomrule
\end{tabular*}
\end{table*}

\subsection{EEF Track Coordinate Alignment}
\label{app:eef_alignment}
We align EEF supervision to the current-view VGGT~\citep{vggt} point map during
training. The full model and both history-ablation variants use this target
preparation. Camera parameters and metric scene depth are recovered from
the training dataset, while EEF positions come from the recorded metric
states. These quantities define auxiliary supervision; action inference
uses RGB observations, robot state, and latent memory.

\paragraph{Joint-state demonstrations.}
When a demonstration provides only joint states, we recover the EEF pose
with forward kinematics (FK) using the robot's kinematic model, such as
its URDF. Let $\boldsymbol{\theta}^{(a)}_{t+k}$ denote the recorded joint
angles of arm $a$ in physical units. Its pose in the common reference
frame is
\begin{equation}
 {}^{w}T_{e_a}(t+k)
 ={}^{w}T_b(t+k)\,\mathrm{FK}_a(\boldsymbol{\theta}^{(a)}_{t+k}),
 \label{eq:eef_fk}
\end{equation}
where $\mathrm{FK}_a$ maps from the robot base frame to the chosen
end-effector frame, including the fixed tool transform when required.
The known base-to-reference transform ${}^{w}T_b$ places the pose in the
same frame used by camera calibration. We take the translation component
at each future demonstration timestamp to form $\mathbf{p}^{w}_{t+k}$.
For dual-arm robots, each kinematic chain is evaluated separately and
both trajectories are expressed in this common frame. These waypoints
then follow the same camera transformation, scale calibration, and
normalization as directly recorded EEF positions.

\paragraph{Replay calibration.}
Each converted demonstration is matched to its raw simulator episode by
its retained action sequence, accounting for no-op removal and gripper
encoding. The episode is replayed to recover the agent-view camera
extrinsics and metric scene depth at each recorded observation. Replayed
EEF positions are checked against the recorded states at every frame.
Image orientation is selected from horizontal/vertical flip combinations
by comparison with the dataset RGB, and checked at the first, middle, and
last frames. The replay acceptance thresholds are $0.05$\,m for the maximum
per-coordinate EEF discrepancy and $20$ for RGB mean absolute error
on the $0$--$255$ intensity scale after resizing to $64\times64$. Per-episode
errors are recorded, and a source fingerprint prevents using caches with
different state/action sequences. Each cached frame contains a
world-to-image-camera transform and metric depths at a regular
$16\times16$ grid of pixel centers.

\paragraph{Camera frame and teacher scale.}
After converting camera conventions, alignment models the teacher's
current-view frame and the calibrated camera frame as differing by a
single scale.
Let $\mathbf{p}^{w}_{t+k}$ be a metric demonstration EEF waypoint and
$[R_t\mid\mathbf{b}_t]$ the cached transform at the current observation.
The transform converts the simulator camera convention and incorporates
the image flips used by the dataset. All future waypoints use this same
current-view transform:
\begin{equation}
 \mathbf{p}^{c}_{t+k}=R_t\mathbf{p}^{w}_{t+k}+\mathbf{b}_t.
 \label{eq:eef_camera_alignment}
\end{equation}
The frozen teacher processes only the current RGB image.
At each cached scene-surface anchor $j$, bilinear sampling of
the raw teacher point map provides a depth $z^{v}_{t,j}$ paired with the
metric simulator depth $d^{c}_{t,j}$. The teacher-units-per-meter scale is
\begin{equation}
 s_t=\operatorname{median}_{j\in\mathcal{V}_t}
 \frac{z^{v}_{t,j}}{d^{c}_{t,j}},
 \label{eq:eef_teacher_scale}
\end{equation}
where $\mathcal{V}_t$ retains finite anchors with
$0.05<d^{c}_{t,j}<5.0$\,m and positive depth ratios.
Rotation and translation come from simulator calibration;
only the scalar depth ratio is estimated from the teacher output.

\paragraph{Shared scene normalization.}
Let $\mathbf{q}_{t,i}$ denote the pooled raw current-view teacher points,
with centroid $\boldsymbol{\mu}_t$ and RMS radius $r_t$:
\begin{equation}
 \begin{aligned}
 \boldsymbol{\mu}_t&=\frac{1}{P^2}\sum_i\mathbf{q}_{t,i},\\
 r_t&=\sqrt{\frac{1}{P^2}\sum_i
       \|\mathbf{q}_{t,i}-\boldsymbol{\mu}_t\|_2^2}.
 \end{aligned}
\end{equation}
The aligned scene and EEF targets are
\begin{equation}
 \begin{aligned}
 \widehat{\mathbf{q}}_{t,i}
   &=\frac{\mathbf{q}_{t,i}-\boldsymbol{\mu}_t}{r_t},\\
 \widehat{\mathbf{p}}_{t+k}
   &=\frac{s_t\mathbf{p}^{c}_{t+k}-\boldsymbol{\mu}_t}{r_t}.
 \end{aligned}
 \label{eq:eef_shared_normalization}
\end{equation}
The implementation floors $r_t$ at $10^{-6}$. Both target segments use
the same scene-derived centroid and radius, preserving the EEF track's
placement relative to the scene.

\paragraph{Validity checks.}
At least 16 usable anchors are required. The median relative deviation
$\operatorname{median}_{j\in\mathcal{V}_t}
|z^{v}_{t,j}/(d^{c}_{t,j}s_t)-1|$ must not exceed $0.25$, and the scale,
extrinsics, and EEF positions must be finite. Samples failing these checks
receive zero EEF-loss weight, while the action and enabled scene losses
remain active.

\begin{figure*}[p]
  \centering
  \includegraphics[width=\textwidth]{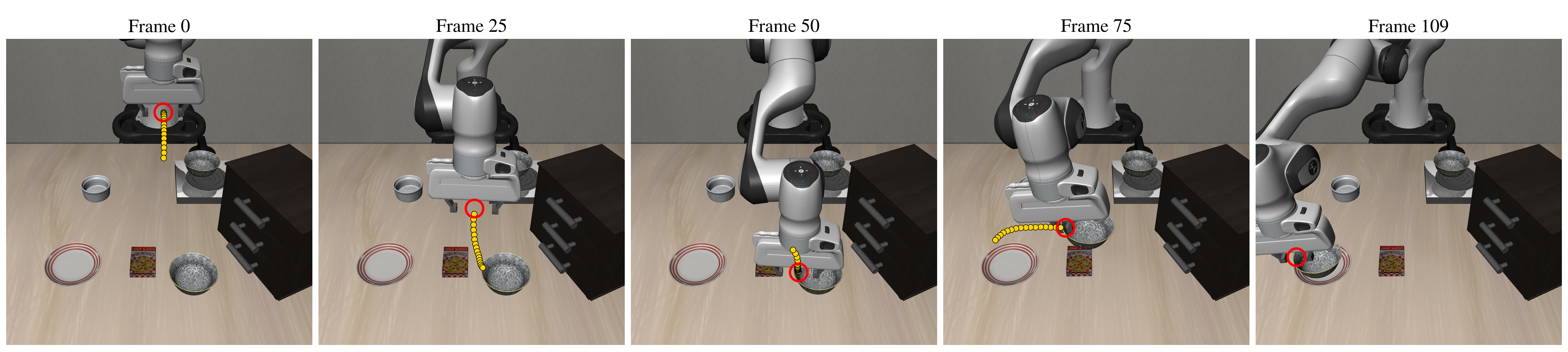}
  \caption{\textbf{Image-plane EEF trajectory alignment over time.}
  Left to right: frames 0, 25, 50, 75, and 100 of the bowl-placement
  sequence. Red circles mark the projected current EEF positions, and
  yellow dots show the projected trajectory segments. The overlays
  visualize the EEF path relative to the gripper and manipulated object
  at successive stages of the demonstration.}
  \label{fig:eef_alignment_projection}
  \par\medskip
  \includegraphics[width=\textwidth]{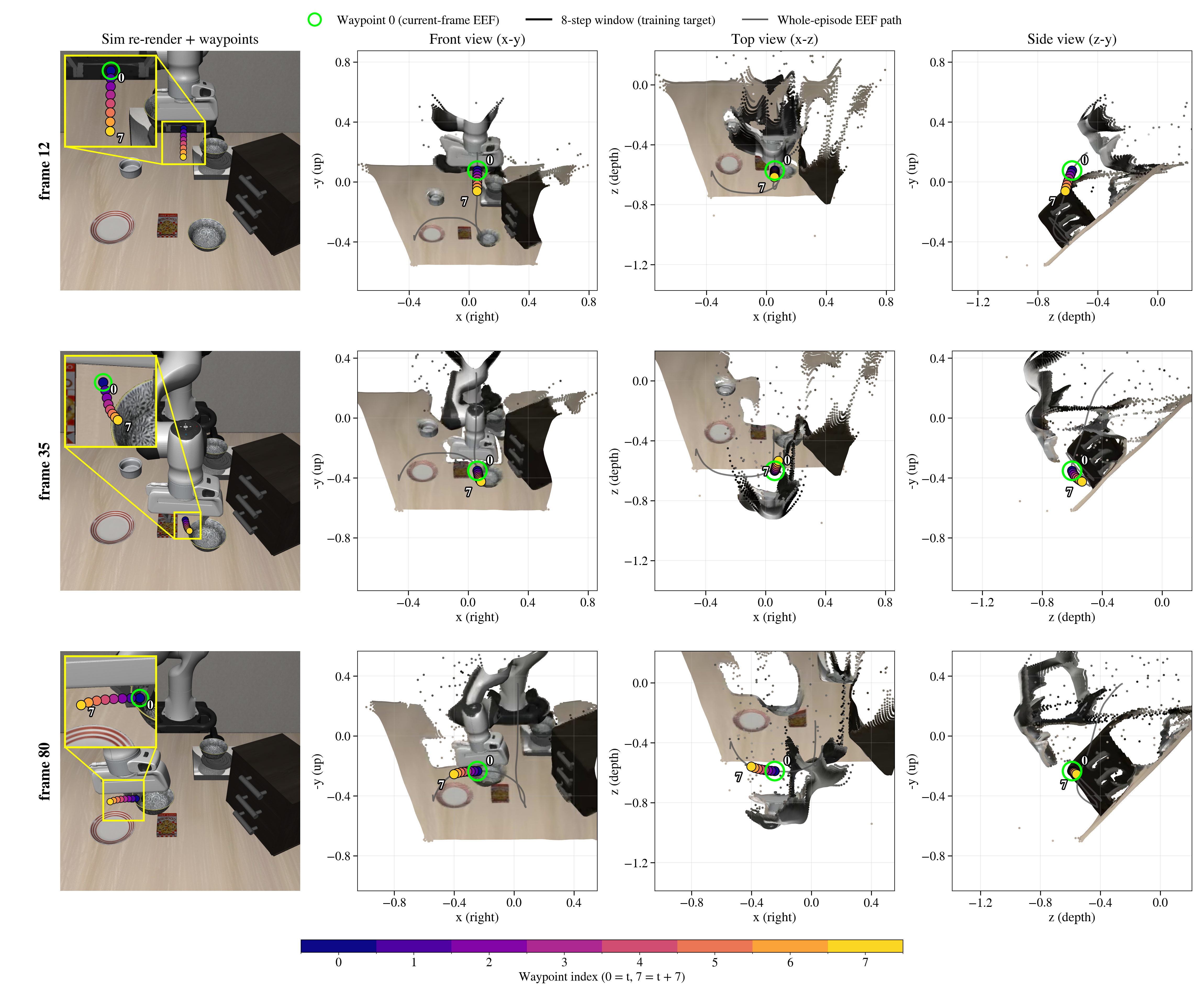}
  \caption{\textbf{EEF trajectory alignment in the shared scene frame.}
  Rows show frames 12, 35, and 80 of a demonstration for picking up the
  black bowl next to the cookie box and placing it on the plate.
  Columns show the RGB image with projected waypoints, followed by front,
  top, and side views of the scene point cloud and aligned EEF trajectory.
  Purple-to-yellow dots denote the eight ordered waypoints; green rings
  mark the current EEF position in the point-cloud views. Black lines
  connect the eight-waypoint segment, and gray lines show the whole-episode
  EEF path in the same current-view normalized frame.}
  \label{fig:eef_alignment_multiview}
\end{figure*}

\paragraph{Qualitative alignment examples.}
\cref{fig:eef_alignment_multiview,fig:eef_alignment_projection} visualize
the alignment pipeline on a bowl-placement demonstration.
At frames 12, 35, and 80, the aligned eight-waypoint segments are shown
alongside the scene point cloud from front, top, and side views
(\cref{fig:eef_alignment_multiview}). The views show how the shared
coordinate frame locates the EEF path relative to the gripper, tabletop,
and nearby objects as the arm approaches and transports the bowl.
\cref{fig:eef_alignment_projection} complements these spatial views with
image-plane projections at five timestamps, showing the trajectory's
placement throughout the manipulation sequence. Together, the two
visualizations illustrate the spatial correspondence between the
demonstration EEF targets and the scene geometry used for supervision.

\section{Geometric Memory Fusion}
\label{app:geo_fusion}
Geometric memory fusion can combine geometric memory and observation, providing action experts with the means to generate motion while avoiding dimensional discrepancies. Let $\widetilde{\mathbf{p}}_t
=\widetilde{\mathbf{x}}^{\mathrm{obs}_1}_t$.
The visual tokens query memory-augmented geometry through
$\mathbf{u}_t=\operatorname{CrossAttn}(
\widetilde{\mathbf{p}}_t,\widetilde{\gtok}_t,
\widetilde{\gtok}_t)$, followed by
$\mathbf{h}_t=\mathbf{u}_t+
\operatorname{FFN}(\operatorname{LN}(\mathbf{u}_t))$.
A channel-wise gate combines the two:
\begin{equation}
  \begin{aligned}
  \mathbf{g}_t&=\sigma(W_g[\mathbf{h}_t;
  \widetilde{\mathbf{p}}_t]+\mathbf{b}_g),\\
  \mathbf{p}'_t&=\mathbf{g}_t\odot\widetilde{\mathbf{p}}_t
  +(1-\mathbf{g}_t)\odot\mathbf{h}_t.
  \end{aligned}
  \label{eq:geo_fusion}
\end{equation}
Cross-attention projects geometry to $D_v$; $W_g$ and
$\mathbf{b}_g$ are learned, while $\sigma$ and $\odot$ denote sigmoid
and elementwise product. The geometric path is zero-initialized and
the gate initially favors visual tokens. The action expert conditions
on $\mathbf{p}'_t$, retrieved $\mathrm{obs}_2$, and robot state.

\section{Per-Task RoboTwin Results}
\label{app:robotwin_full}
\begin{table*}[t]
\caption{\textbf{Per-task results on RoboTwin 2.0.} Success rate (\%) for all
50 tasks under Clean (C) and Randomized (R) settings, 100 episodes per task.
\methodname is compared against its two ablated variants: without the geometry
branch (w/o Geo) and without temporal memory (w/o Mem).}
\label{tab:robotwin_full}
\begin{center}
\tiny
\setlength{\tabcolsep}{2.5pt}
\resizebox{\textwidth}{!}{%
\begin{tabular}{@{}rl cc cc cc @{\hspace{6pt}} rl cc cc cc@{}}
\toprule
& & \multicolumn{2}{c}{Full} & \multicolumn{2}{c}{w/o Geo} & \multicolumn{2}{c}{w/o Mem}
& & & \multicolumn{2}{c}{Full} & \multicolumn{2}{c}{w/o Geo} & \multicolumn{2}{c}{w/o Mem} \\
\cmidrule(lr){3-4}\cmidrule(lr){5-6}\cmidrule(lr){7-8}
\cmidrule(lr){11-12}\cmidrule(lr){13-14}\cmidrule(lr){15-16}
\# & Task & C & R & C & R & C & R & \# & Task & C & R & C & R & C & R \\
\midrule
1 & adjust\_bottle & 100 & 100 & 100 & 100 & 100 & 100 & 26 & place\_can\_basket & 87 & 93 & 79 & 79 & 74 & 76 \\
2 & beat\_block\_hammer & 99 & 94 & 93 & 93 & 98 & 92 & 27 & place\_cans\_plasticbox & 99 & 100 & 100 & 99 & 98 & 97 \\
3 & blocks\_ranking\_rgb & 99 & 95 & 97 & 96 & 99 & 99 & 28 & place\_container\_plate & 100 & 100 & 98 & 98 & 99 & 98 \\
4 & blocks\_ranking\_size & 87 & 86 & 86 & 85 & 83 & 80 & 29 & place\_dual\_shoes & 85 & 81 & 84 & 70 & 87 & 88 \\
5 & click\_alarmclock & 92 & 97 & 87 & 93 & 91 & 95 & 30 & place\_empty\_cup & 100 & 100 & 100 & 100 & 100 & 100 \\
6 & click\_bell & 97 & 95 & 91 & 95 & 97 & 97 & 31 & place\_fan & 98 & 100 & 97 & 97 & 96 & 97 \\
7 & dump\_bin\_bigbin & 97 & 98 & 96 & 97 & 96 & 95 & 32 & place\_mouse\_pad & 90 & 92 & 81 & 83 & 83 & 88 \\
8 & grab\_roller & 98 & 98 & 97 & 98 & 100 & 100 & 33 & place\_object\_basket & 83 & 86 & 87 & 80 & 83 & 87 \\
9 & handover\_block & 93 & 92 & 89 & 86 & 96 & 94 & 34 & place\_object\_scale & 92 & 92 & 90 & 92 & 90 & 88 \\
10 & handover\_mic & 100 & 98 & 97 & 100 & 97 & 96 & 35 & place\_object\_stand & 97 & 94 & 99 & 95 & 96 & 91 \\
11 & hanging\_mug & 44 & 40 & 41 & 42 & 38 & 30 & 36 & place\_phone\_stand & 90 & 100 & 90 & 96 & 86 & 95 \\
12 & lift\_pot & 100 & 100 & 100 & 100 & 100 & 100 & 37 & place\_shoe & 99 & 97 & 100 & 100 & 98 & 98 \\
13 & move\_can\_pot & 87 & 91 & 86 & 90 & 89 & 92 & 38 & press\_stapler & 100 & 100 & 100 & 99 & 100 & 98 \\
14 & move\_pillbottle\_pad & 100 & 98 & 99 & 98 & 99 & 100 & 39 & put\_bottles\_dustbin & 84 & 88 & 87 & 90 & 87 & 94 \\
15 & move\_playingcard\_away & 100 & 99 & 100 & 98 & 99 & 98 & 40 & put\_object\_cabinet & 88 & 85 & 87 & 87 & 86 & 89 \\
16 & move\_stapler\_pad & 78 & 68 & 75 & 67 & 63 & 74 & 41 & rotate\_qrcode & 87 & 89 & 86 & 87 & 89 & 88 \\
17 & open\_laptop & 99 & 100 & 99 & 100 & 100 & 100 & 42 & scan\_object & 95 & 95 & 93 & 94 & 93 & 95 \\
18 & open\_microwave & 99 & 99 & 93 & 97 & 89 & 89 & 43 & shake\_bottle & 100 & 100 & 100 & 100 & 100 & 100 \\
19 & pick\_diverse\_bottles & 73 & 72 & 72 & 72 & 74 & 79 & 44 & shake\_bottle\_horizontally & 100 & 100 & 100 & 99 & 100 & 99 \\
20 & pick\_dual\_bottles & 84 & 89 & 83 & 79 & 83 & 89 & 45 & stack\_blocks\_three & 81 & 84 & 83 & 82 & 85 & 87 \\
21 & place\_a2b\_left & 80 & 82 & 79 & 76 & 80 & 80 & 46 & stack\_blocks\_two & 99 & 99 & 100 & 99 & 100 & 100 \\
22 & place\_a2b\_right & 75 & 72 & 77 & 71 & 80 & 73 & 47 & stack\_bowls\_three & 91 & 91 & 93 & 92 & 89 & 86 \\
23 & place\_bread\_basket & 97 & 93 & 92 & 95 & 94 & 94 & 48 & stack\_bowls\_two & 99 & 99 & 99 & 96 & 100 & 97 \\
24 & place\_bread\_skillet & 89 & 89 & 87 & 84 & 85 & 84 & 49 & stamp\_seal & 86 & 88 & 77 & 93 & 77 & 89 \\
25 & place\_burger\_fries & 95 & 99 & 96 & 97 & 100 & 98 & 50 & turn\_switch & 63 & 65 & 62 & 71 & 66 & 73 \\
\midrule
\multicolumn{8}{r}{} & \multicolumn{2}{l}{\textbf{Average (50 tasks)}}
& \multicolumn{2}{c}{\textbf{91.10 / 91.24}}
& \multicolumn{2}{c}{\textbf{89.68 / 89.74}}
& \multicolumn{2}{c}{\textbf{89.84 / 90.52}} \\
\bottomrule
\end{tabular}}
\end{center}
\end{table*}

The per-task RoboTwin 2.0~\citep{robotwin} results are as reported in \cref{tab:robotwin_full}. All three variants use task-adaptive execution horizons. Their displayed
means are recomputed from the 50 per-task scores for each setting.
By analyzing the actual settings of different tasks and failure cases, we can determine which type of information a task is more sensitive to.

\section{Case-Study Details}
\label{app:cases}
\paragraph{Data and configuration.}
The examples use episode~0 from two tasks in each of the Clean and Randomized
RoboTwin datasets.
We decode point maps and EEF tracks at four selected frames per episode
using 20 Euler steps and a fixed noise seed.

Memory visualization separately enables streaming inference, resets all banks
before each episode, and supplies only preceding sampled observations to
retrieval.

\paragraph{Metric definition and coverage.}
For corresponding point coordinates, we compute
\begin{equation}
 e_{\mathrm{3D}} = \sqrt{\frac{1}{N}\sum_{i=1}^{N}
       \|\widehat{\mathbf{p}}_i-\mathbf{p}_i\|_2^2}.
 \label{eq:case_rmse}
\end{equation}
Scene error uses $N=256$ patch correspondences in centroid-centered,
unit-RMS-radius coordinates. EEF error uses $N=100$ waypoints across both
arms and is converted from meters to millimeters. Scene coordinates
and EEF displacements are co-registered. \Cref{tab:case_coverage} averages
the four per-frame RMSE values within each case.
Across all 16 frames, per-frame point-map RMSE ranges from $0.032$ to
$0.152$ in normalized units.

\begin{table*}[t]
\caption{\textbf{Coverage of the 200K reconstruction diagnostics.}
All four available cases and all 16 reconstructed frames are included.
Point-map errors are unweighted; EEF errors include moving and stationary arms.}
\label{tab:case_coverage}
\centering\small
\begin{tabular}{llcc}
\toprule
Task & Setting & Point-map RMSE & EEF RMSE (mm) \\
\midrule
Open microwave & Clean & 0.076 & 3.69 \\
Open microwave & Randomized & 0.094 & 3.74 \\
Stack three blocks & Clean & 0.070 & 3.07 \\
Stack three blocks & Randomized & 0.064 & 3.64 \\
\bottomrule
\end{tabular}
\end{table*}

\begin{figure*}[tp]
  \centering
  \includegraphics[width=\textwidth]{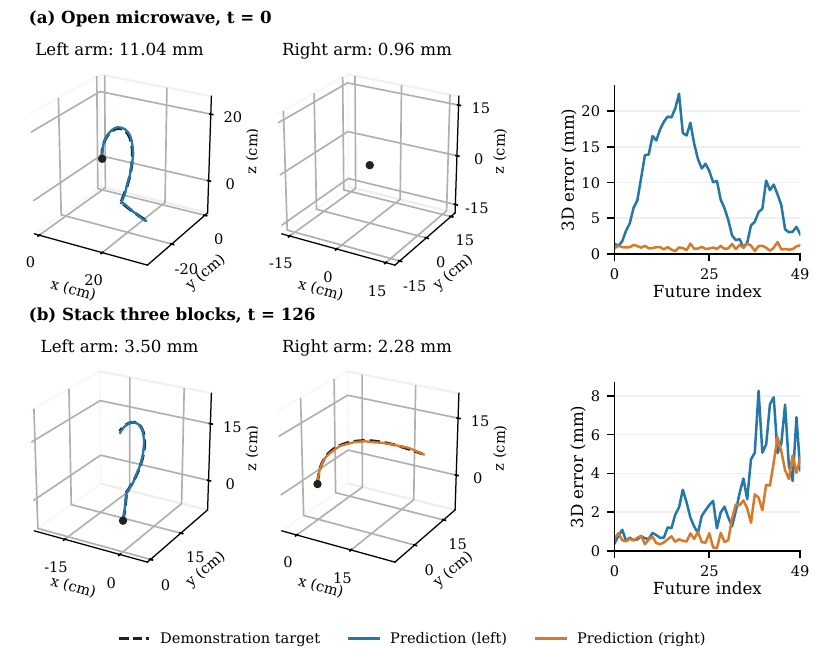}
  \caption{\textbf{Detailed future EEF tracks for the examples in
  \cref{fig:case_geometry}.} Dashed curves are demonstration-derived targets;
  solid curves are predictions from the auxiliary head. Each arm contributes
  50 waypoints. Dots mark target starting positions. Both arms within a row
  use equal spatial scale, with coordinates displayed in centimeters. The right panels show
  Euclidean error at every future index.}
  \label{fig:case_eef}
\end{figure*}

\begin{figure*}[tp]
    \centering
    \includegraphics[width=\linewidth,height=0.90\textheight,keepaspectratio]{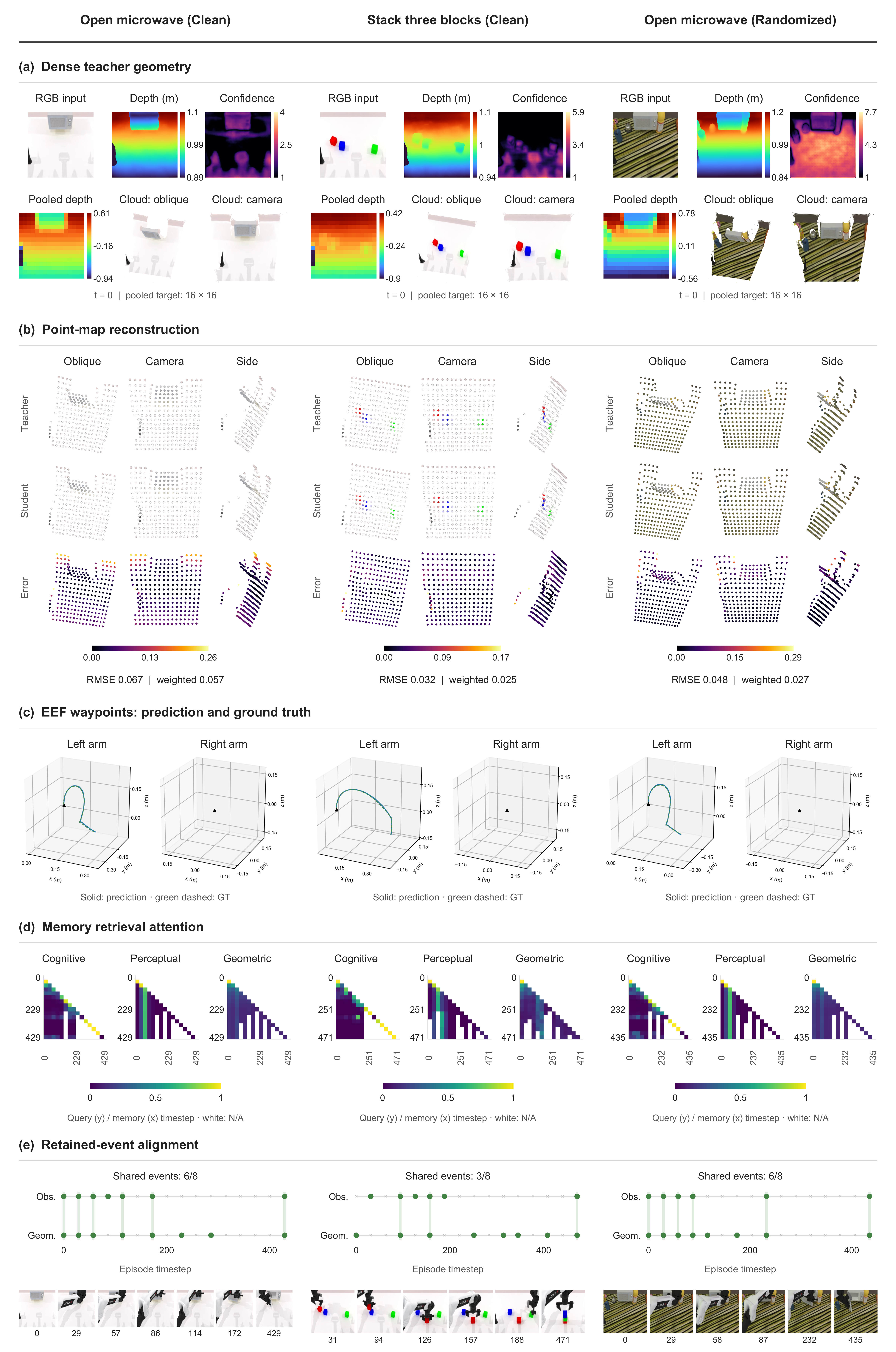}
    \caption{
\textbf{Geometry reconstruction and memory analysis on RoboTwin manipulation tasks.}
For each example, we visualize:
(a) dense geometry targets provided by the frozen VGGT teacher,
(b) reconstructed point maps from GeoPlan representations,
(c) predicted future end-effector tracks,
(d) retrieval attention across perceptual, cognitive, and geometric memories,
and (e) retained-event alignment between observation and geometric memory banks.
}
    \label{fig:robotwincases}
\end{figure*}

\paragraph{Reading the memory visualization.}
The two observational banks consolidate independently. The observational
row in \cref{fig:case_geometry} plots the intersection of their stored
timestamps, while the geometric row plots its own bank. A cross in the
observational row means that at least one observational bank does not retain
that timestamp.
In the Randomized stacking replay, the 16 writes cause eight merges per
capacity-eight bank. The observational banks share six final timestamps;
four of these, 94, 126, 314, and 471, also remain in geometric memory.
Additional reconstruction and memory diagnostics are shown in \Cref{fig:robotwincases}

\section{Real World Experiments Details}

\subsection{Hardware and Experimental Setup}

We conduct real-world experiments on two dual-arm robot platforms, Astribot S1 and AgileX Cobot Magic 2, as shown in \Cref{fig:RealWorldExper}. These platforms provide different robot embodiments for evaluating the applicability of \methodname in real-world manipulation scenarios.

The Astribot S1 is a dual-arm humanoid manipulation platform equipped with two 7-DoF arms and dexterous end-effectors. The robot is equipped with a three-camera visual system, including one head-mounted RGB camera and two wrist-mounted RGB cameras, which provide global scene observations and close-range views for manipulation.

The AgileX Cobot Magic 2 is a dual-arm collaborative robot platform equipped with two 6-DoF robotic arms and parallel grippers. Similar to Astribot S1, it uses a three-camera configuration consisting of one head-mounted RGB camera and two wrist-mounted RGB cameras.

\subsection{Task Description}

We design five real-world manipulation tasks on two robot platforms to evaluate different aspects of \methodname, including spatial understanding, object-level manipulation, and long-horizon task execution. The tasks cover diverse interaction patterns, ranging from precise object placement to sequential manipulation with changing task states.

\begin{figure*}[p]
    \centering
    \includegraphics[width=1.0\linewidth]{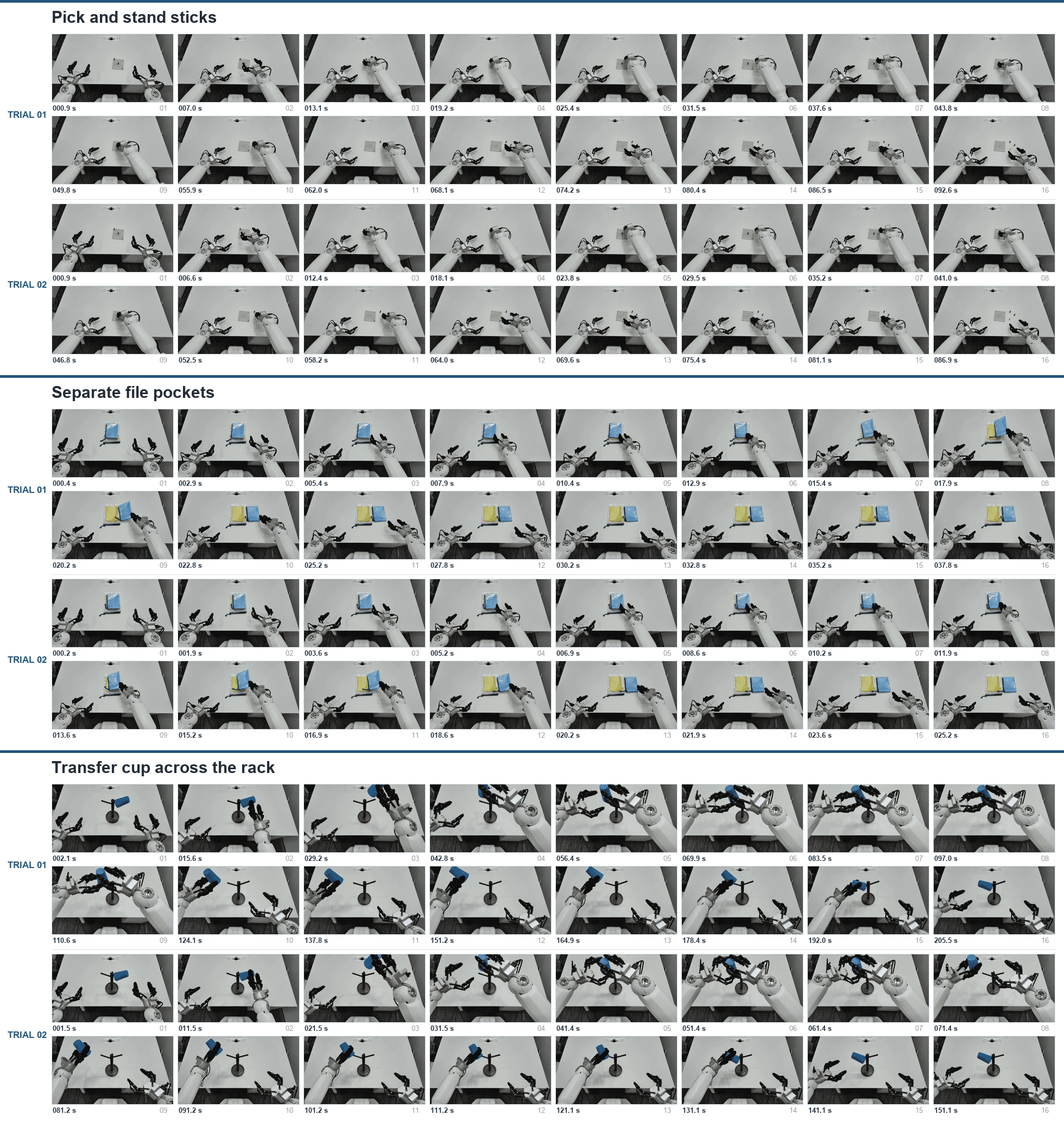}
    \caption{
\textbf{Successful real-world rollouts on Astribot S1.}
The figure presents representative execution sequences for three manipulation tasks:
pick and stand sticks, separate file pockets, and transfer cup across the rack.
GeoMem-VLA completes long-horizon manipulation procedures under randomized object configurations,
demonstrating accurate spatial interaction and effective task-state retention.
}
    \label{fig:astribotrolloutsuccess}
\end{figure*}

\paragraph{Astribot S1 Tasks.} Real-world rollouts are shown in \Cref{fig:astribotrolloutsuccess}.

\textit{Pick and stand sticks.}
In this task, three colored sticks (blue, red, and yellow) are placed on a pegboard. The robot is required to sequentially pick up each stick and place it upright on the table on the right side according to the specified order.

\textit{Transfer a cup across the rack.}
A blue cup is placed on the right side of a wooden rack. The robot first grasps the cup with the right gripper, removes it from the rack, transfers it to the left gripper in the air, and finally places it on the left side of the rack.

\textit{Separate file pockets.}
Three file pockets with different colors and sizes are stacked on the table. The robot needs to identify the target blue file pocket, grasp it with the right gripper, and place it at the designated location on the right side.

\begin{figure*}[p]
    \centering
    \includegraphics[width=1.0\linewidth]{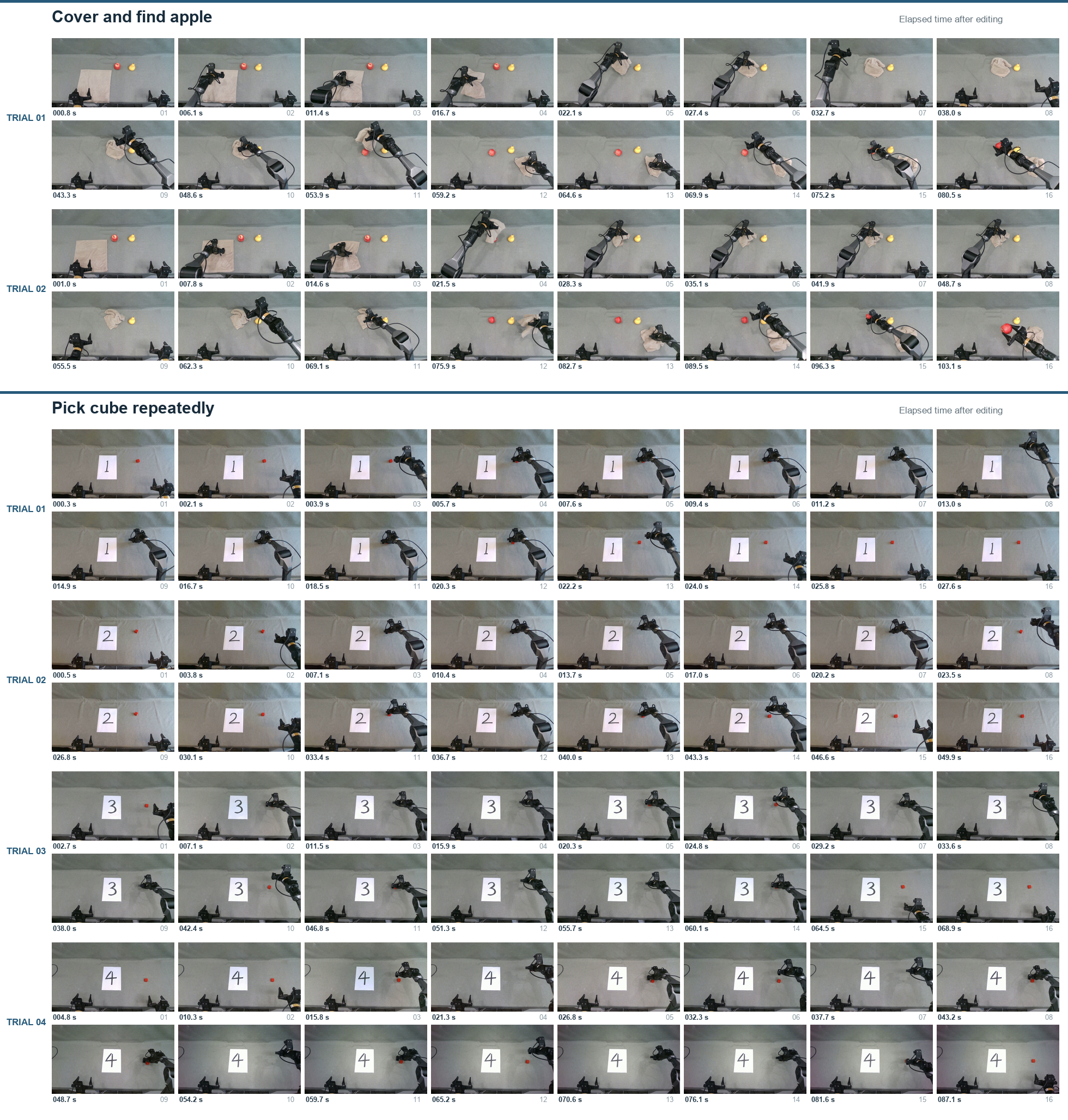}
    \caption{\textbf{Successful real-world rollouts on AgileX Cobot Magic 2.} The figure presents representative execution sequences for two manipulation tasks: cover and find apple under towel and pick cube repeatedly. GeoMem-VLA completes long-horizon manipulation procedures involving hidden-object reasoning and repeated interactions, demonstrating effective spatial understanding and task-state retention.}
    \label{fig:agilex_success}
\end{figure*}

\paragraph{AgileX Cobot Magic 2 Tasks.}
Real-world rollouts are shown in \Cref{fig:agilex_success}.

\textit{Pick n times cube.}
A number instruction is provided on the table, indicating the required number of repetitions. The robot needs to repeatedly grasp and place the red cube on the right side of the table for exactly $n$ times.

\textit{Cover and find the apple under towel.}
An apple, a pear, and a towel are placed on the table. The robot first grasps the towel and covers the apple, then removes the towel to reveal the hidden apple, and finally grasps the apple.

\subsection{Evaluation Protocol}

For each real-world task, we conduct 10 independent evaluation rollouts under randomized initial configurations. Before each rollout, object positions are randomly initialized within the demonstrated workspace, while task-specific parameters (e.g., the target number in the \textit{pick n times cube} task) are randomly assigned.

During evaluation, the robot executes the policy in a closed-loop manner without external intervention. A rollout is considered successful only when the robot completes the entire task procedure and satisfies the predefined task objective. Partial completion of a task is counted as failure.

The success rate is calculated as:

\begin{equation}
SR = \frac{N_{\mathrm{success}}}{N_{\mathrm{total}}} \times 100\%.
\end{equation}

The reported success rate for each task is computed over the 10 evaluation rollouts.

\subsection{Baseline Implementation}

We compare \methodname with three representative VLA baselines:
LingBot-VLA 2.0~\citep{lingbotvla2},
$\pi_{0.5}$~\citep{pi05}, and
OpenVLA-OFT~\citep{openvlaoft}. To ensure a fair comparison, all baselines are fine-tuned and evaluated following their official configurations and recommended training pipelines.

For each task, the baseline models are fine-tuned using the same task-specific demonstration data collected in our experiments. The adapted models are then evaluated under the same rollout settings, including identical initial configurations, evaluation episodes, and success criteria.

\subsection{Additional Qualitative Results}

\begin{figure*}[tp]
    \centering
    \includegraphics[width=\linewidth,height=0.90\textheight,keepaspectratio]{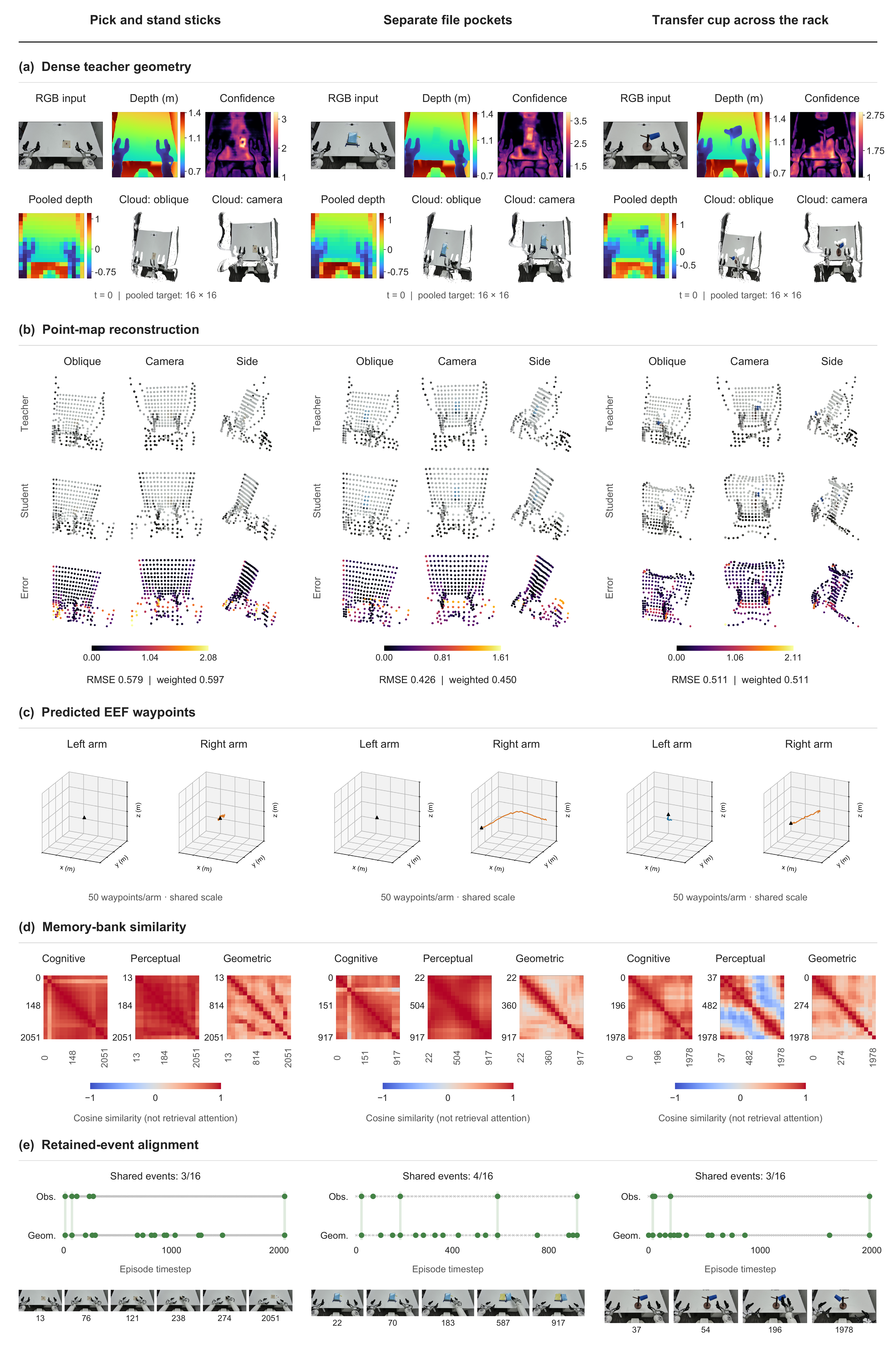}
    \caption{
\textbf{Geometry reconstruction and memory analysis on real-world Astribot tasks.}
For three representative tasks, we visualize:
(a) dense teacher geometry estimated from RGB observations,
(b) reconstructed point maps from GeoPlan tokens,
(c) predicted future end-effector waypoints,
(d) memory retrieval similarity across different representation streams,
and (e) retained-event alignment between observation and geometric memories.
}
    \label{fig:astribotcases}
\end{figure*}

\begin{figure*}[tp]
    \centering
    \includegraphics[width=\linewidth,height=0.90\textheight,keepaspectratio]{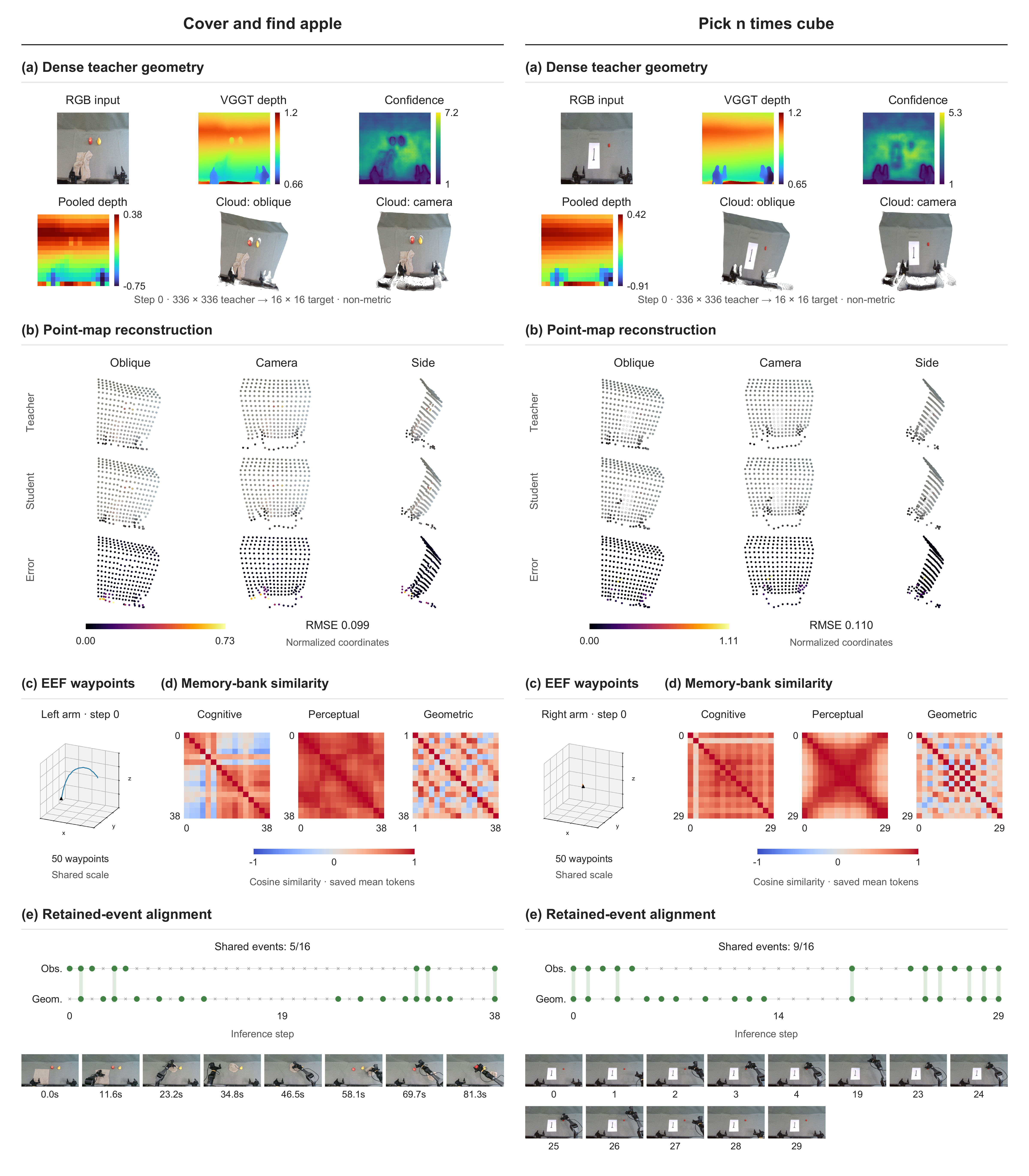}
    \caption{
\textbf{Geometry reconstruction and memory analysis on real-world AgileX tasks.}
For two representative tasks, we visualize:
(a) dense teacher geometry estimated from RGB observations,
(b) reconstructed point maps from GeoPlan tokens,
(c) predicted future end-effector waypoints,
(d) memory retrieval similarity across different representation streams,
and (e) retained-event alignment between observation and geometric memories.
}
    \label{fig:agilexcases}
\end{figure*}

We provide additional details of real-world experiments as shown in \Cref{fig:astribotcases} and \Cref{fig:agilexcases} to further illustrate the effectiveness of \methodname in spatially precise manipulation and long-horizon task execution.

\end{document}